\documentclass[10pt,compsoc]{IEEEtran}

\usepackage{amsmath,amssymb,float,graphicx,algorithm2e,subcaption,bm, multirow,array}
\usepackage{epsf,epsfig,pstricks,pst-node,color}

\RequirePackage{graphicx}
\DeclareMathOperator*{\argmax}{arg\,max}

\newcommand{\vecx}{\mathbf{x}}

\newcommand{\graph}{\ensuremath{\mathcal{G}}}
\newcommand{\pdfBN}{\ensuremath{p^{\mathcal{B}}}}

\newcommand{\matX}{\ensuremath{\mathbf{X}}}
\newcommand{\BN}{\ensuremath{\mathcal{B}}}

\def\reg{{\rm\ooalign{\hfil
     \raise.07ex\hbox{\scriptsize R}\hfil\crcr\mathhexbox20D}}}

\renewenvironment{IEEEbiography}[1]
  {\IEEEbiographynophoto{#1}}
  {\endIEEEbiographynophoto}

\begin{document}

\title{Efficient and Robust Machine Learning for Real-World Systems} 

\author{Franz~Pernkopf,~\IEEEmembership{Senior Member,~IEEE}, Wolfgang~Roth, Matthias Z\"ohrer, Lukas Pfeifenberger, G\"unther Schindler, Holger Fr\"oning, Sebastian~Tschiatschek, Robert Peharz, Matthew Mattina, Zoubin Ghahramani
\IEEEcompsocitemizethanks{\IEEEcompsocthanksitem 
W.~Roth, M.~Z\"ohrer, L.~Pfeifenberger, and F.~Pernkopf are with the Department of Electrical Engineering, Laboratory of Signal Processing and Speech Communication, Graz University of Technology, Austria.\protect\\
G. Schindler and H. Fr\"oning are with the Institute of Computer Engineering, Ruperts Karls University, Heidelberg, Germany.\protect\\
S.~Tschiatschek is with Microsoft Research, Cambridge, UK.\protect\\
R.~Peharz is with the Machine Learning Group, Department of Engineering, University of Cambridge, UK.\protect\\
Z.~Ghahramani is with the Machine Learning Group, Department of Engineering, University of Cambridge, UK and Uber AI Labs, California, USA\protect\\
M.~Mattina is with Arm Research, Arm Ltd., Cambridge, UK.
\protect\\
Correspondence e-mail: pernkopf@tugraz.at}
\thanks{}}

\IEEEcompsoctitleabstractindextext{%
\begin{abstract}
While machine learning is traditionally a resource intensive task, embedded systems, autonomous navigation and the vision of the Internet-of-Things fuel the interest in resource efficient approaches.
These approaches require a carefully chosen trade-off between performance and resource consumption in terms of computation and energy.
On top of this, it is crucial to treat uncertainty in a consistent manner in all but the simplest applications of machine learning systems. 
In particular, a desideratum for any real-world system is to be robust in the presence of outliers and corrupted data, as well as being ``aware'' of its limits, i.e.\ the system should maintain and provide an uncertainty estimate over its own predictions.
These complex demands are among the major challenges in current machine learning research and key to ensure a smooth transition of machine learning technology into every day's applications.
In this article, we provide an overview of the current state of the art of machine learning techniques facilitating these real-world requirements.
First we provide a comprehensive review of resource-efficiency in deep neural networks with focus on techniques for model size reduction, compression and reduced precision.
These techniques can be applied during training or as post-processing and are widely used  to reduce both computational complexity and memory footprint.
As most (practical) neural networks are limited in their ways to treat uncertainty, we contrast them with probabilistic graphical models, which readily serve these desiderata by means of probabilistic inference.
In that way, we provide an extensive overview of the current state-of-the-art of robust and efficient machine learning for real-world systems.
\end{abstract}

}

\maketitle    

\IEEEdisplaynotcompsoctitleabstractindextext

\IEEEpeerreviewmaketitle

\section{Introduction}

Machine learning is a key technology in the 21st century and the main contributing factor for many recent performance boosts in computer vision, natural language processing, speech recognition and signal processing. 
Today, the main application domain and comfort zone of machine learning applications is the ``virtual world'', as found in recommender systems, stock market prediction, and social media services.
However, we are currently witnessing a transition of machine learning moving into ``the wild'', where most prominent examples are autonomous navigation for personal transport and delivery services, and the Internet of Things (IoT).
Evidently, this trend opens several real-world challenges for machine learning engineers.

\begin{figure}[t]
\begin{center}
 \includegraphics[width=6cm]{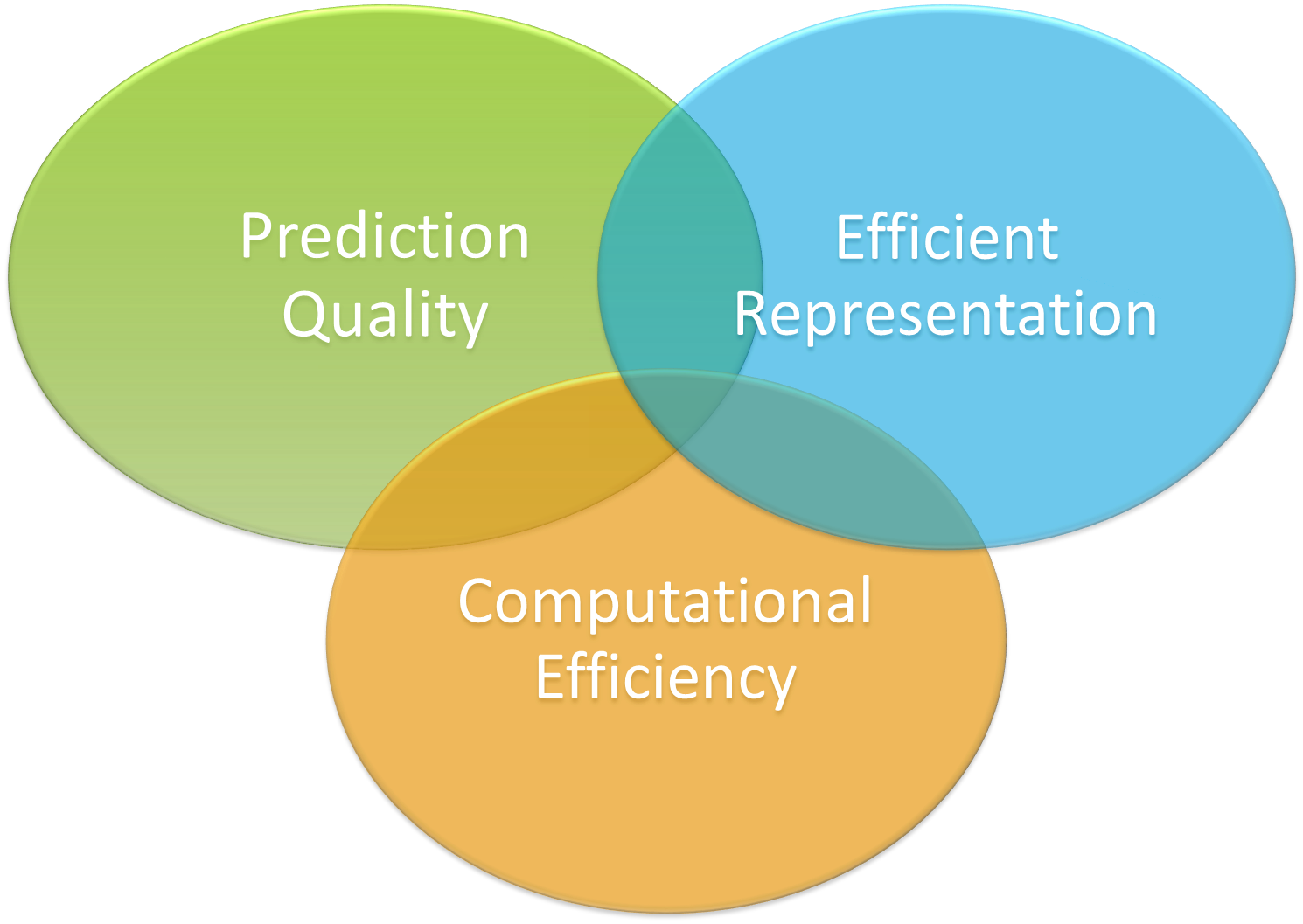}
\caption{Aspects of resource-efficient machine learning models.}
 \label{EffClass}
\end{center}
\end{figure}
Current machine learning approaches prove particularly effective when big amounts of data and ample computing resources are available.
However, in real-world applications the computing infrastructure during the operation phase is typically limited, which effectively rules out most of the current resource-hungry machine learning approaches.
There are several key challenges -- illustrated in Figure~\ref{EffClass} -- which have to be jointly considered to facilitate machine learning in real-world applications:
\begin{itemize} 
\item Efficient representation: The model complexity, i.e. the number of model parameters, should match the usually limited resources in deployed systems, in particular regarding memory footprint.
\item Computational efficiency: The machine learning model should be computationally efficient during inference, exploiting the available hardware optimally with respect to time and energy. For instance, power constraints are key for autonomous and embedded systems, as the device lifetime for a given battery charge needs to be maximized, or constraints set by energy harvesters need to be met.
\item Prediction quality: The focus of classical machine learning is mostly on optimizing the prediction quality of the models. For embedded devices, model complexity versus prediction quality trade-offs must be considered to achieve good prediction performance while simultaneously reducing computational complexity and memory requirements.
\end{itemize}

Furthermore, in the ``non-virtual'' world, we have only limited control over data quality.
Corrupted data, missing inputs, and outliers are the rule rather than the exception.
These real-world conditions require a high degree of robustness of machine learning systems under corrupted inputs, as well as the model's ability to deliver well-calibrated predictive uncertainty estimates.
This point is especially crucial if the system shall be involved in any critical decision making processes.

In this article, we review the state of the art in machine learning with regard to these real-world requirements.
We first focus on deep neural networks (DNN), the currently predominant machine learning models.
While being the driving factor behind many recent success stories, DNNs are notoriously data and resource hungry,
a property which has recently renewed significant research interest in resource-efficient approaches.
The first part of this tutorial is dedicated to an extensive overview of these approaches, all of which exploit the following two generic strategies to
(i) reduce model size in terms of number of weights and/or neurons, or 
(ii) reduce arithmetic precision of parameters and/or computational units.
Evidently, these two basic techniques are almost ``orthogonal directions'' towards efficiency in DNNs, and they can be naturally combined, e.g.~one can both sparsify a model and reduce arithmetic precision.

Nevertheless, most research emphasizes one of these two techniques, so that we discuss them separately.
In Section \ref{sec:model_size_reduction_DNNs}, we first discuss approaches to reduce the model size in DNNs, using pruning techniques, weight sharing, factorized representations, and knowledge distillation.
In Section \ref{sec:reduced_precision_DNNs}, we focus on techniques for reduced arithmetic precision in DNNs.
When driven to the extreme, this approach leads to discrete DNNs, with only a few values for weights and/or activations.
Even reducing precision down to binary or ternary values works reasonably well and essentially reduces DNNs to hardware-friendly logical circuits.
This extreme reduction, however, introduces challenging discrete optimization problems.
Besides various optimization heuristics, such as the straight-through estimator, we also discuss an alternative approach, casting the problem as Bayesian posterior inference.
The latter approach can be naturally tackled with variational approximations, leading to a continuous optimization problem.

The Bayesian approach also readily incorporates uncertainty treatment into machine learning models, as weight uncertainty represented by the (approximate) posterior directly translates into well-calibrated output uncertainties.
Clearly, however, processing a full posterior during test time is computationally demanding and largely opposed to our primary goal of resource-efficiency.
Thus, in practice a compromise must be made, realizable via small ensembles of discrete DNNs.

As an alternative to DNNs we discuss classical probabilistic graphical models (PGMs) in Section~\ref{sec:pgms}.
PGMs naturally lend themselves towards resource-efficient machine learning systems, typically yielding models which are several orders of magnitude smaller than DNNs, while still obtaining decent predictive performance.
Furthermore, they treat uncertainty in a natural way by virtue of statistical inference and often dramatically outperform DNNs when a considerable number of input features are missing.
Moreover, generative or hybrid learning approaches yield well-calibrated uncertainties over both inputs and outputs which can naturally be exploited in outlier and abnormality detection.
Similarly to DNNs, PGMs can be subjected to efficiency optimizations by employing structure learning and reduced-precision parameters.
In particular, for inference scenarios like classification, they can be highly efficient, requiring only integer additions.

In Section~\ref{sec:experiments} we substantiate our discussion with experimental results.
First, we exemplify the trade-off between execution time, memory footprint and predictive accuracy in DNNs on a CIFAR-10 classification task. 
Subsequently, we provide an extensive comparison of various hardware-efficient strategies for DNNs, using the challenging task of ImageNet classification.
In particular, this overview shows that sensible trade-offs can be achieved with very low numeric precision, such as only one bit per activation and DNN weight.
We demonstrate that these trade-offs can be readily exploited on today's hardware, by benchmarking the core operation of binary DNNs (BNNs), i.e.~binary matrix multiplication, on NVIDIA Tesla K80 and ARM Cortex-A57 architectures.
Furthermore, a complete real-world signal processing example using BNNs is discussed in Section~\ref{sec:speech_mask}.
In this example we develop a complete speech enhancement system employing an efficient BNN-based speech mask estimator, which shows negligible performance degradation while allowing memory savings of factor 32 and speed-ups of roughly a factor 10.
Furthermore, exemplary results comparing PGMs and DNNs on the classical MNIST data set are provided where the focus is on prediction performance and number of bits necessary for representing the models. DNNs slightly outperform PGMs on MNIST while PGMs are able to naturally handle missing feature scenarios.
An example of randomly missing features during model testing is finally provided.

\section{Deep Neural Networks}

DNNs are the currently dominant approach in machine learning, and have led to significant performance boosts in various application domains, such as computer vision~\cite{Krizhevsky2012}, speech and natural language processing~\cite{Hinton2012, Sutskever2014}.   
In~\cite{BengioCV13}, key aspects of deep models have been identified explaining some of the performance gains, namely, the re-use of features in consecutive layers and the degree of abstraction of features at higher layers. 

Furthermore, the performance improvements can be largely attributed to increasing hardware capabilities that enabled the training of ever-increasing network architectures and the usage of big data. 
Since recently there is growing interest in making DNNs available for embedded devices by developing fast and energy-efficient architectures with little memory requirements. 
These methods reduce either the number of connections and parameters (Section~\ref{sec:model_size_reduction_DNNs}), the parameters' precision (Section~\ref{sec:reduced_precision_DNNs}), or both, as discussed in the sequel.

\subsection{Model Size Reduction in DNNs}    \label{sec:model_size_reduction_DNNs}

In the following, we review methods that reduce the number of weights and neurons in DNNs using techniques like pruning, sharing, but also more complex methods like knowledge distillation and special data structures. 

\subsubsection{Weight Pruning and Neuron Pruning}
One of the earliest approaches to reduce network size is LeCun et al.'s \emph{optimal brain damage} algorithm \cite{LeCun1989}.
Their main finding is that pruning based on weight magnitude is suboptimal and they propose a pruning scheme based on the increase in loss function.
Assuming a pre-trained network, a local second-order Taylor expansion with a diagonal Hessian approximation is employed that allows to estimate the change in loss function caused by weight pruning without re-evaluating the costly network function.
Removing parameters is alternated with re-training the pruned network.
In that way, the model can be reduced significantly without deteriorating its performance.
Hassibi and Stork \cite{Hassibi1992} found the diagonal Hessian approximation to be too restrictive, and their \emph{optimal brain surgeon} algorithm uses an approximated full covariance matrix instead.
While their method, similarly as \cite{LeCun1989}, prunes weights that cause the least increase in loss function, the remaining weights are simultaneously adapted to compensate for the negative effect of weight pruning.
This bypasses the need to alternate several times between pruning and re-training the pruned network.

However, it is not clear whether these approaches scale up to modern DNN architectures since computing the required (diagonal) Hessians is substantially more demanding (if not intractable) for millions of weights.
Therefore, many of the more recently proposed techniques still resort to magnitude based pruning.
Han et al.\ \cite{Han2015} alternate between pruning connections below a certain magnitude threshold and re-training the pruned network.
The results of this simple strategy are impressive, as the number of parameters in pruned networks is an order of magnitude smaller (9$\times$ for AlexNet and $13\times$ for VGG-16) than in the original networks.
Hence, this work shows that neural networks are in general heavily over-parametrized.
In a follow-up paper, Han et al.\ \cite{Han2016} proposed \emph{deep compression}, which extends the work in \cite{Han2015} by a parameter quantization and parameter sharing step, followed by Huffman coding to exploit the non-uniform weight distribution.
This approach yields a $35$-$49\times$ improvement in memory footprint and consequently  a reduction in energy consumption of $3$-$5\times$.

Guo et al.\ \cite{Guo2016} discovered that irreversible pruning decisions limit the achievable sparsity and that it is useful to reincorporate weights pruned in an earlier stage.
In addition to full weight matrices, they maintain a set of weight masks that determine whether a weight is currently pruned or not.
Their method alternates between updating the weights based on gradient descent, and updating the weight masks by thresholding.
Most importantly, weight updates are also applied to weights that are currently pruned such that pruned weights can reappear if their value exceeds a certain threshold.
This yields a $17.7\times$ parameter reduction for AlexNet without deteriorating performance.

Wen et al.\ \cite{Wen2016} incorporated group lasso regularizers in the objective to obtain different kinds of sparsity in the course of training.
They were able to remove filters, channels, and even entire layers for architectures where shortcut connections are used.

In \cite{Graves2011,Blundell2015}, variational inference is employed to train for each connection a weight variance $w_{\sigma^2}$ in addition to a single (mean) weight $w_{\mu}$.
After training, weights are pruned according to the ``signal-to-noise ratio'' $|w_{\mu}/w_{\sigma}|$.
Molchanov et al.\ \cite{Molchanov2017} proposed a method based on Kingma et al.'s variational dropout \cite{Kingma2015} which interprets dropout as performing variational inference with specific prior and approximate posterior distributions.
Within this framework, the otherwise fixed dropout rates appear as free parameters that can be optimized to improve a variational lower bound.
In \cite{Molchanov2017}, this freedom is exploited to optimize weight dropout rates such that weights can be safely pruned if their dropout rate is close to one.
This idea has been extended in \cite{Louizos2017} by using sparsity enforcing priors and assigning dropout rates to groups of weights that are all connected to the same neuron which in turn allows the pruning of entire neurons.
Furthermore, they show how their approach can be used to determine an appropriate bit width for each weight by exploiting the well-known connection between Bayesian inference and the minimum description length (MDL) principle \cite{Gruenwald2007}.
We elaborate more on Bayesian approaches in Section \ref{sec:quantization_bayesian_approaches}.

In \cite{Mariet2016}, a determinantal point process (DPP) is used to find a group of neurons that are diverse and exhibit little redundancy.
Conceptionally, a DPP for a given ground set $\mathcal{S}$ defines a distribution over subsets $S \subseteq \mathcal{S}$ where subsets containing diverse elements have high probability.
The DPP is used to sample a diverse set of neurons and the remaining neurons are then pruned.
To compensate for the negative effect of pruning, the outgoing weights of the kept neurons are adapted so as to minimize the activation change of the next layer.

\subsubsection{Weight Sharing}
A further technique to reduce the model size is weight-sharing.
In~\cite{Chen2015}, a hashing function is used to randomly group network connections into ``buckets'', where the connections in each bucket shares a single weight value.
This has the advantage that weight assignments need not be stored explicitly but are given implicitly by the hashing function.
This allows to train $10\times$ smaller networks while the predictive performance is essentially unaffected.
Ullrich et al.\ \cite{Ullrich2017} extended the soft weight-sharing approach proposed in \cite{Nowlan1992} to achieve both weight sharing and sparsity.
The idea is to select a Gaussian mixture model prior over the weights and to train both the weights as well as the parameters of the mixture components.
During training, the mixture components collapse to point measures and each weight gets attracted by a certain weight component.
After training, weight sharing is obtained by assigning each weight to the mean of the component that best explains it, and weight pruning is obtained by fixing the mean of one component to zero and assigning it a relatively high mixture mass.

\subsubsection{Knowledge Distillation}
\emph{Knowledge distillation} \cite{Hinton2015} is an indirect approach where first a large model (or an ensemble of models) is trained, and subsequently soft-labels obtained from the large model are used as training data for a smaller model.
The smaller models achieve performances almost identical to that of the larger models which is attributed to the valuable information contained in the soft-labels.
Inspired by knowledge distillation, Korattikara et al.\ \cite{Korattikara2015} reduced a large ensemble of DNNs, used for obtaining Monte-Carlo estimates of a posterior predictive distribution, to a single DNN.

\subsubsection{Special Weight Matrix Structures}
There also exist approaches that aim at reducing the model size on a more global scale by (i) reducing the parameters required to represent the large matrices involved in DNN computations, or by (ii) employing certain matrix structures that facilitate low-resource computation in the first place.
Denil et al.\ \cite{Denil2013} propose to represent weight matrices $\mathbf{W} \in \mathbb{R}^{m \times n}$ using a low-rank approximation $\mathbf{UV}$ with $\mathbf{U} \in \mathbb{R}^{m \times k}$, $\mathbf{V} \in \mathbb{R}^{k \times n}$, and $k < \min\{m,n\}$ to reduce the number of parameters.
Instead of learning both factors $\mathbf{U}$ and $\mathbf{V}$, prior knowledge, such as smoothness of pixel intensities in an image, is incorporated to compute a fixed $\mathbf{U}$ using kernel-techniques or auto-encoders, and only the factor $\mathbf{V}$ is learned.
This approach is motivated by training only a subset of the weights and predicting the values of the other weights from this subset.
In \cite{Novikov2015}, the Tensor Train matrix format is employed to substantially reduce the number of parameters required to represent large weight matrices of fully-connected layers.
Their approach enables the training of very large fully-connected layers with relatively few parameters and they show better performance than simple low-rank approximations.
Denton et al.\ \cite{Denton2014} propose specific low-rank approximations and clustering techniques for individual layers of pre-trained convolutional DNNs (CNN) to both reduce memory-footprint and computational overhead.
Their approach yields substantial improvements for both the computational bottleneck in the convolutional layers and the memory bottleneck in the fully-connected layers.
By fine-tuning after applying their approximations, the performance degradation is kept at a decent level.
Jaderberg et al.\ \cite{Jaderberg2014} propose two different methods to approximate pre-trained CNN filters as combinations of rank-1 basis filters to speed-up computation.
The rank-1 basis filters are obtained either by minimizing a reconstruction error of the original filters or by minimizing a reconstruction error of the outputs of the convolutional layers.
Lebedev et al.\ \cite{Lebedev2015} approximate the convolution tensor by a low-rank approximation using non-linear least squares.
Subsequently, the convolution using this low-rank approximation is performed by four consecutive convolutions, each with a smaller filter, to reduce the computation time substantially.
In \cite{Cheng2015}, the weight matrices of fully-connected layers are restricted to circulant matrices $\mathbf{W} \in \mathbb{R}^{n \times n}$, which are fully specified by only $n$ parameters.
While this dramatically reduces the memory footprint of fully-connected layers, circulant matrices also facilitate faster computation as matrix-vector multiplication can be efficiently computed using the fast Fourier transform.
In a similar vein, Yang et al.\ \cite{Yang2015} reparameterize matrices $\mathbf{W} \in \mathbb{R}^{n \times n}$ of fully-connected layers using the Fastfood transform as $\mathbf{W}{=}\mathbf{SHG \Pi HB}$, where $\mathbf{S}$, $\mathbf{G}$, and $\mathbf{B}$ are diagonal matrices, $\mathbf{\Pi}$ is a random permutation matrix, and $\mathbf{H}$ is the Walsh-Hadamard matrix.
This reparameterization requires only a total of $4 n$ parameters, and similar as in \cite{Cheng2015}, the fast Hadamard transform enables the efficient computation of matrix-vector products.
Iandola et al.\ \cite{Iandola2016} introduced \emph{SqueezeNet}, a special CNN structure that requires far less parameters while maintaining similar performance as AlexNet on the ImageNet data set.
Their structure incorporates both $1 \times 1$ and $3 \times 3$ convolutions, and they use, similar as in \cite{Lin2014}, global average pooling of per-class feature maps that are directly fed into the softmax in order to avoid fully-connected layers that typically consume the most memory.
Furthermore, they show that their approach is compatible with deep compression \cite{Han2016} to reduce the memory footprint to less than $0.5$MB.

\subsection{Reduced Precision in DNNs}    \label{sec:reduced_precision_DNNs}
As already mentioned before, the two main approaches to reduce the model size of DNNs are structure sparsification and reducing parameter precision.
These approaches are to some extent orthogonal techniques to each other.
Both strategies reduce the memory footprint accordingly and are vital for the deployment of DNNs in many real-world applications.
Importantly, as pointed out in \cite{Han2016,Chen2016,Horowitz2014}, reduced memory requirements are the main contributing factor to reduce the energy consumption as well.
Furthermore, model sparsification also impacts the computational demand measured in terms of number of arithmetic operations.
Unfortunately, this reduction in the mere number of arithmetic operations usually does not directly translate into savings of wall-clock time, as current hardware and software are not well-designed to exploit model sparseness \cite{Zhang2016}.
Reducing parameter precision, on the other hand, proves very effective for improving execution time \cite{Vanhoucke2011}.
When the latter point is driven to the extreme, i.e. assuming binary weights $w \in \{-1,1\}$ or ternary weights $w \in \{-1,0,1\}$ in conjunction with binary inputs and/or hidden units $x \in \{-1,1\}$, floating or fixed point multiplications are replaced by hardware-friendly logical XNOR and bitcount operations.
In that way, a sophisticated DNN is essentially reduced to a logical circuit.
However, training such discrete-valued DNNs\footnote{Due to finite precision, in fact any DNN is discrete valued. However, we use this term here to highlight the extremely low number of values.} is delicate as they cannot be directly optimized using gradient based methods.
In the sequel, we provide a literature overview of approaches that use reduced-precision computation to facilitate low-ressource training and/or testing.

\subsubsection{Stochastic Rounding}
Approaches for reduced-precision computations date back at least to the early 1990s.
H{\"{o}}hfeld and Fahlman \cite{Hoehfeld1992a,Hoehfeld1992b} rounded the weights during training to fixed-point format with different numbers of bits.
They observed that training eventually stalls, as small gradient updates are always rounded to zero.
As a remedy, they proposed stochastic rounding, i.e.~rounding values to the nearest value with a probability proportional to the distance to the nearest value.
These quantized gradient updates are correct in expectation, do not cause training to stall, and yield substantially less bits than when using deterministic rounding.

More recently, Gupta et al.~\cite{Gupta2015} have shown that stochastic rounding can also be applied for modern deep architectures, as demonstrated on a hardware prototype.
Courbariaux et al.~\cite{Courbariaux2015a} empirically studied the effect of different numeric formats (floating point, fixed point, and dynamic fixed point) with varying bit widths on the performance of DNNs. 
Lin et al.~\cite{Lin2015} proposes a method to dramatically reduce the number of multiplications required during training.
At forward propagation, the weights are stochastically quantized to either binary weights $w \in \{-1,1\}$ or ternary weights $w \in \{-1,0,1\}$ to remove the need for multiplications at all.
During backpropagation, inputs and hidden neurons are quantized to powers of two, reducing multiplications to cheaper bit-shift operations, leaving only a negligible number of floating-point multiplications to be computed.
However, the speed-up is limited to training since for testing the full-precision weights are required.
Lin et al.\ \cite{Lin2016} consider fixed-point quantization of pre-trained full-precision DNNs.
They formulate an optimization problem that minimizes the total number of bits required to store the weights and the activations under the constraint that the total output signal-to-quantization noise ratio is larger than a certain pre-specified value.
A closed-form solution of the convex objective yields layer-specific bit widths.

\subsubsection{Straight-Through Estimator (STE)}
\begin{figure}[t]
\begin{center}
 \includegraphics[width=0.5\textwidth]{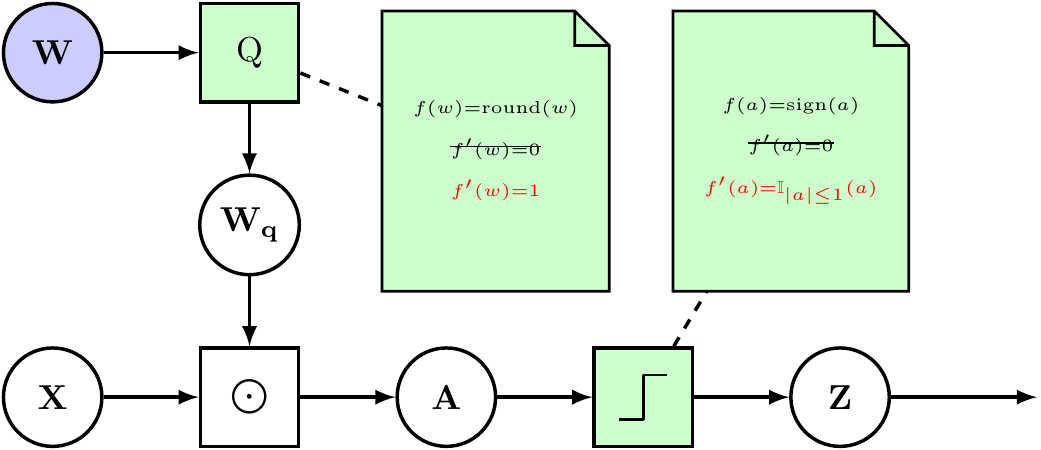}
\caption{A simplified building block of a DNN using the straight-through estimator (STE).
Real-valued weights $\mathbf{W}$ are quantized with a quantization function $Q$ to obtain quantized weights $\mathbf{W_q}$.
These quantized weights are combined with the inputs $\mathbf{X}$ via $\odot$ (e.g. convolution) to obtain activations $\mathbf{A}$ which are subsequently fed into a non-linear activation function to obtain $\mathbf{Z}$.
The blue circle indicates the real-valued parameters $\mathbf{W}$ that should be updated with gradient descent, but due to the green boxes the gradient $\nabla_{\mathbf{W}}$ is zero.
During backpropagation when the chain-rule is invoked, the STE replaces the zero-derivative by a non-zero surrogate derivative to allow gradient information to flow backwards.
}
 \label{fig:ste}
\end{center}
\end{figure}

In recent years, the straight-through estimator (STE) \cite{Bengio2013} became the method of choice for training DNNs with weights that are represented using a very small number of bits.
Quantization operations, being piecewise constant functions with either undefined or zero gradients, are not applicable to gradient-based learning using backpropagation.
The idea of the STE is to simply replace piecewise constant functions with a non-zero artificial derivative during backpropagation, as illustrated in Figure~\ref{fig:ste}.
This allows gradient information to flow backwards through piecewise constant functions to subsequently update parameters based on the approximated gradients.
Note that this also allows for the training of DNNs with quantized activation functions such as the sign function.
Approaches based on the STE typically maintain a set of full-precision weights that are quantized during forward propagation.
After backpropagation, gradient updates are applied to the full-precision weights.
At test time, the full-precision weights are abandoned and only the quantized reduced-precision weights are kept.

In \cite{Courbariaux2015b}, binary-weight DNNs are trained using STE to get rid of expensive floating-point multiplications.
They consider deterministic and stochastic rounding during forward propagation and update a set of full-precision weights based on the gradients of the quantized weights.
In \cite{Hubara2016}, the STE is used to quantize both the weights and the activations to a single bit with sign functions. This reduces the computational burden dramatically as floating-point multiplications and additions are reduced to hardware-friendly logical XNOR and bitcount operations, respectively.
Li et al.\ \cite{Li2016} trained ternary weights $w \in \{-a,0,a\}$ by setting weights lower than a certain threshold $\Delta$ to zero, and setting weights to either $-a$ or $a$ otherwise.
Their approach determines $a>0$ and $\Delta$ by approximately minimizing a $\ell^2$-norm between a full-precision matrix and the quantized ternary weight matrix.

Zhu et al.\ \cite{Zhu2017} extended their work to ternary weights $w \in \{-a,0,b\}$ by learning the factors $a>0$ and $b>0$ using gradient updates and a different threshold $\Delta$ based on the maximum full-precision weight magnitude for each layer.
These asymmetric weights considerably improve performance compared to symmetric weights as used in \cite{Li2016}.

Rastegari et al.\ \cite{Rastegari2016} approximate full-precision weight filters in CNNs as $\mathbf{W} = \alpha \mathbf{B}$ where $\alpha$ is a scalar and $\mathbf{B}$ is a binary weight matrix.
This reduces the bulk of floating-point multiplications inside the convolutions to additions or subtractions, and only requires a single multiplication per output neuron with the scalar $\alpha$.
In a further step, the layer inputs are quantized in a similar way to perform the convolution with only efficient XNOR operations and bitcount operations, followed by two floating-point multiplications per output neuron.
For backpropagation, the STE is used.
Lin et al.\ \cite{Lin2017} generalized the ideas of \cite{Rastegari2016} and approximate the full-precision weights with linear combinations of multiple binary weight filters for improved classification accuracy.
Motivated by the fact that weights and activations typically exhibit a non-uniform distribution, Miyashita et al.\ \cite{Miyashita2016} proposed to quantize values to powers of two.
Their representation allows to get rid of expensive multiplications and they report higher robustness to quantization than linear rounding schemes using the same number of bits.
While activation binarization methods using the sign function can be seen as approximating commonly used sigmoid functions such as tanh, Cai et al.\ \cite{Cai2017} proposed a half-wave Gaussian quantization that more closely resembles the predominant ReLU activation function.
Benoit et al. \cite{Benoit2018} proposed a quantization scheme that accurately approximates floating point operations using integer arithmetic only to speed-up computation.
During training, their forward pass simulates the quantization step to keep the performance of the quantized DNN close to the performance when using single-precision.
At test time, weights are represented as 8-bit integer values, reducing the memory footprint by a factor of four.

Zhou et al.\ \cite{Zhou2016} presented several quantization schemes that allow for flexible bit widths, both for weights and activations.
Furthermore, they also propose a quantization scheme for backpropagation to facilitate low-resource training, and, in agreement with earlier work mentioned above, they note that stochastic quantization is essential for their approach.
In \cite{Wu2018}, weights, activations, weight gradients, and activation gradients are subject to customized quantization schemes that allow for variable bit widths, and that facilitate integer arithmetic during training and testing.
In contrast to \cite{Zhou2016}, the work in \cite{Wu2018} accumulates weight changes to low-precision weights instead of full-precision weights. 
While most work on quantization based approaches is empirical, some recent work gained more theoretical insights \cite{Li2017,Anderson2018}. 

\subsubsection{Bayesian Approaches}   \label{sec:quantization_bayesian_approaches}
\begin{figure}[t]
 \centering
 \includegraphics[width=7.5cm]{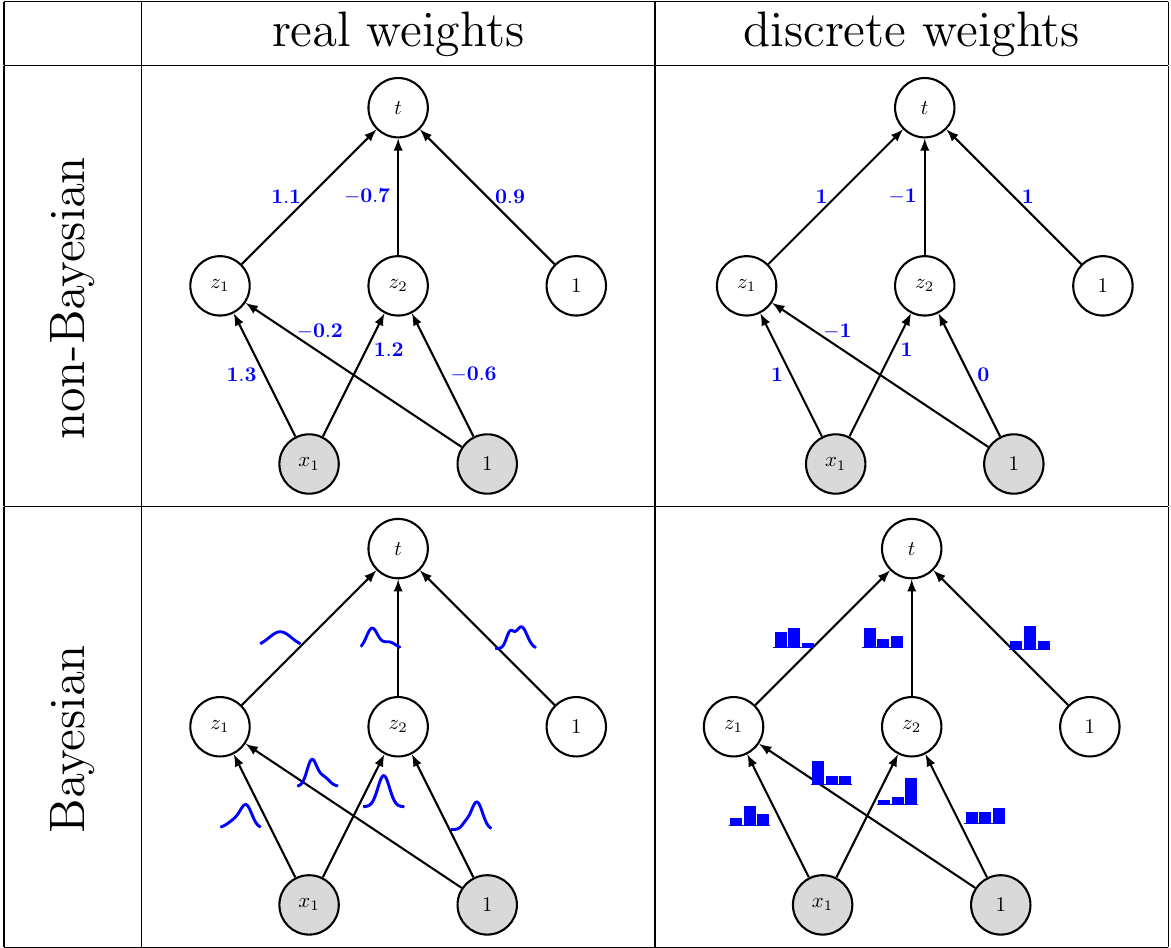}
 \caption{Overview of real-valued vs.\ discrete-valued NNs and Bayesian vs.\ non-Bayesian NNs. The aim is to obtain a single discrete-valued NN (top right) with a good performance. In the Bayesian approach this is achieved by training a distribution over discrete-valued NNs (bottom right) and subsequently deriving a single discrete-valued NN from that distribution.}
 \label{fig:bayesian_discrete_overview}
\end{figure}

Alternatively to approaches reviewed so far, there exist approaches to train reduced-precision DNNs without any quantization at all.
A particularly attractive option for learning discrete-valued DNNs are Bayesian approaches where a distribution over the weights is maintained instead of a fixed weight assignment.
This is illustrated in Figure \ref{fig:bayesian_discrete_overview}.
Given a prior $p(\mathbf{W})$ on the weights, a data set $\mathcal{D}$, and a likelihood $p(\mathcal{D}|\mathbf{W})$ that is defined by a DNN, we can use Bayes' rule to infer a posterior distribution over the weights, i.e.
\begin{align}
 p(\mathbf{W}|\mathcal{D}) = \frac{p(\mathcal{D}|\mathbf{W}) \ p(\mathbf{W})}{p(\mathcal{D})} \propto p(\mathcal{D}|\mathbf{W}) \ p(\mathbf{W}). \label{eq:bayes_rule}
\end{align}
From a Bayesian viewpoint, training DNNs can be seen as seeking a point of maximum probability within such a posterior distribution.
However, this approach is problematic for discrete-valued DNNs since gradient-based optimization cannot be applied.
A solution is to approximate the full posterior $p(\mathbf{W}|\mathcal{D})$ using a tractable variational distribution $q(\mathbf{W})$, and to subsequently use either a maximum of $q(\mathbf{W})$ or to sample an ensemble from it.
A common approximation to $q(\mathbf{W})$ assumes independence among the weights -- known as mean-field assumption -- which renders variational inference tractable.
To compute a maximum of $q(\mathbf{W})$ or to sample an ensemble of discrete weight sets is straightforward under this assumption.

Soudry et al.\ \cite{Soudry2014} approximate the true posterior $p(\mathbf{W}|\mathcal{D})$ using expectation propagation \cite{Minka2001} in an online fashion with closed-form updates.
Starting with an uninformative approximation $q(\mathbf{W})$, their approach combines the current approximation $q(\mathbf{W})$ (serving as the prior in \eqref{eq:bayes_rule}) with the likelihood for a data set $\mathcal{D} = \{(\mathbf{x},c)\}$ comprising only a single sample to obtain a refined posterior.
Since this refinement step is not available in closed-form, they propose several approximations in order to yield a more amenable objective.
A different Bayesian approach for discrete-valued weights has been presented by Roth and Pernkopf \cite{Roth2018}.
They approximate the posterior $p(\mathbf{W}|\mathcal{D})$ by minimizing the Kullback-Leibler (KL)-divergence $KL(q(\mathbf{W})||p(\mathbf{W}|\mathcal{D}))$ with respect to the parameters $\boldsymbol{\nu}$ of the approximation $q(\mathbf{W})$.
The KL-divergence is commonly decomposed as
\begin{align}
 &KL(q(\mathbf{W})||p(\mathbf{W}|\mathcal{D})) = KL(q(\mathbf{W})||p(\mathbf{W})) \notag \\
 &\qquad- \mathbb{E}_{q(\mathbf{W})}[ \log p(\mathcal{D}|\mathbf{W}) ] + \log p(\mathcal{D}). \label{eq:variational_objective}
\end{align}
This expression does not involve the intractable posterior $p(\mathbf{W}|\mathcal{D})$ and the evidence $\log p(\mathcal{D})$ is constant with respect to $\boldsymbol{\nu}$.
The KL term can be seen as a regularizer that pulls the approximate posterior $q(\mathbf{W})$ towards the prior $p(\mathbf{W})$ whereas the expected log-likelihood captures the data.
The expected log-likelihood in \eqref{eq:variational_objective} is intractable, but it can be optimized using the so-called reparameterization trick and stochastic optimization \cite{Rezende2014,Kingma2014,Blundell2015}.
In \cite{Roth2018}, it has been proposed to optimize an approximation to \eqref{eq:variational_objective} using similar techniques as in \cite{Soudry2014}.
Compared to directly optimizing in the discrete weight space, this approach has the advantage that the real-valued parameters $\boldsymbol{\nu}$ of the approximation $q(\mathbf{W})$ can be optimized with gradient-based techniques.
In particular, they trained feed-forward DNNs using sign activations with 3-bit weights in the first layer and ternary weights in the remaining layers and achieved results that are on par with results obtained using the STE \cite{Hubara2016}.

\section{Probabilistic Models}    \label{sec:pgms}

Efficiency, as discussed so far, is clearly one of the main challenges for machine learning systems in real-world applications.
However, in the evolution of machine learning techniques we are facing another challenge which must not be underestimated: uncertainty.
The further we move machine learning towards ``the wild'', the more circumstances get out of our control.
Consequently, any machine learning system which shall realistically be applied in real-world applications, must to some degree facilitate a mechanism to treat uncertainty in all aspects, i.e.~inputs, internal states, the environment and outputs.

To this end, probabilistic methods \cite{Ghahramani2015,Pernkopf2014Introduction} are arguably the method of choice when it comes to reasoning under uncertainty.
The classical probabilistic models are PGMs, which represent variables in data sets as nodes in a graph and direct variable dependencies as edges between the nodes. 
PGMs are a naturally resource-efficient representation of a probability distribution, especially when sparse model structures are used~\cite{Per10}. 
Additionally, their memory footprint can be reduced when considering reduced-precision parameters~\cite{Tsc14a, Tsc15}. 

Compared to DNNs the prediction quality of PGMs is usually inferior. 
Due to this fact, discriminative and hybrid learning paradigms have been proposed to substantially improve the prediction performance. 
On the other hand, probabilistic methods are often significantly more data efficient than deep learning techniques, in particular when rich domain structure can be exploited \cite{Lake2015, George2017}. 
In this article, we focus on learning \emph{directed} PGMs also called \emph{Bayesian networks} (BNs)~\cite{Kol09,Pernkopf2014Introduction}, while efficient implementations of undirected graphical models have been considered in \cite{piatkowski2016integer}. 

\subsection{Learning Bayesian Networks}

A Bayesian network (BN) $\mathcal{B} = (\mathcal{G}, \mathcal{P}_\graph)$ describes the dependencies of a set of random variables $\mathbf{X} = [X_0, \ldots, X_L]$ by a directed graph $\mathcal{G}$, i.e.\ the \emph{structure} of the BN, and a set of conditional probability distributions $\mathcal{P}_\graph$ associated with the nodes of $\mathcal{G}$, i.e.\ the \emph{parameters}.
The $i$\textsuperscript{th} node of $\mathcal{G}$ corresponds to the random variable $X_i$ and the edges in $\mathcal{G}$ encode conditional independence properties between the random variables.
For each $X_i$ there is a conditional probability distribution $p(X_i \mid Pa(X_i))$, where $Pa(X_i)$ denotes the parents of variable $X_i$ according to $\mathcal{G}$.
The BN defines a probability distribution over $\mathbf{X}$ as $\pdfBN(\mathbf{X}) = \prod_{i=0}^L p(X_i \mid Pa(X_i))$.
When using BNs for classification, one of the random variables $\mathbf{X}$ takes the special role of the class variable, yielding a joint distribution $\pdfBN(C,\mathbf{X})$ for $\pdfBN(C,X_1,\ldots,X_L)$.

Both parameters and structure can be learned using either generative or discriminative objectives~\cite{Per10}.
The inference performance (i.e.~the prediction error) of BNs can in general be boosted when discriminative learning paradigms are used.
In the generative approach, we exploit the parameter posterior, yielding the maximum-a-posterior (MAP), maximum likelihood (ML) or the Bayesian approach. 
In discriminative learning, alternative objective functions are considered, such as conditional log-likelihood (CLL)~\cite{Wet03,Roo05,Gre05}, classification rate (CR), or margin~\cite{Guo05,Per11}, which can be applied for both structure learning and for parameter learning~\cite{Per10,Per13}. 
Furthermore, hybrid parameter learning has been proposed unifying the generative and discriminative learning paradigms~\cite{Bou04,Bis07,Peh13,ROTH2018Hybrid}, and combine their respective advantages (allowing, e.g.\ to consistently treat missing data).

\subsubsection{Parameter Learning}

The conditional probability densities (CPDs) of BNs are usually of some parametric form, which can be optimized either generatively or discriminatively. 
Several approaches to optimize BN parameters are discussed in the following.

\begin{itemize}
 \item \textbf{Generative Parameters.} In generative parameter learning the goal is to capture the generative process generating the available data, i.e.~generative parameters are based on the idea of \emph{approximating} the true underlying data distribution with a distribution $\pdfBN(C, \matX)$. 
An example of this paradigm is \emph{maximum-likelihood} learning, i.e.~optimizing the likelihood 
   \begin{align}
     \mathcal{P}_\graph^{\textnormal{ML}} = \arg \max_{\mathcal{P}_\graph} \prod_{n=1}^N \pdfBN(c^{(n)}, \vecx^{(n)}),
   \end{align}
 where $c^{(n)}$ and $\vecx^{(n)}$ are the instantiations of $C$ and $X_1, \ldots, X_L$ for the $n$\textsuperscript{th} training data point respectively.
 Maximum likelihood parameters minimize the KL-divergence between $\pdfBN(C, \matX)$ and empirical data distribution, and thus, under mild assumptions, the KL-divergence between $\pdfBN(C, \matX)$ and the true distribution~\cite{Kol09}.
 \item \textbf{Discriminative Parameters.} In discriminative learning one is interested in parameters yielding good classification performance on new samples from the data distribution. Discriminative learning is especially advantageous when the assumed model distribution $\pdfBN(C, \matX)$ cannot capture the underlying data distribution well, as for example when rather limited BN structures are used~\cite{Roos2005}. 
Several objectives for discriminative parameter learning have been proposed.
Here, we consider the \emph{maximum-conditional-likelihood} (MCL) objective~\cite{Roos2005} and the \emph{maximum-margin} (MM) objective~\cite{Guo05,Per11}.   
MCL parameters are obtained by maximizing
\begin{align}
 \mathcal{P}_\graph^{\textnormal{MCL}} = \arg \max_{\mathcal{P}_\graph} \prod_{n=1}^N \pdfBN(c^{(n)} | \vecx^{(n)}),
\end{align}   
where $\pdfBN(C|\matX)$ denotes the conditional distribution of $C$ given as
\begin{align}
  \pdfBN(C|\matX) = \frac{\pdfBN(C,\matX)}{\pdfBN(\matX)}.
\end{align}
   
MM parameters $\mathcal{P}_\graph^{\textnormal{MM}}$ are obtained by maximizing
\begin{align}
    \mathcal{P}_\graph^{\textnormal{MM}} = \arg \max_{\mathcal{P}_\graph} \prod_{n=1}^N \min\left(\gamma, d^\BN(c^{(n)}, \vecx^{(n)})  \right),
\end{align}    
where $d^\BN(c^{(n)}, \vecx^{(n)})$ is the \emph{probabilistic margin} of the $n$\textsuperscript{th} sample, given as
\begin{align}
 d^\BN(c^{(n)}, \vecx^{(n)}) = \frac{\pdfBN(c^{(n)} | \vecx^{(n)})}{\max_{c \neq c^{(n)}} \pdfBN(c | \vecx^{(n)})}.
\end{align}
The margin can be interpreted as the model's confidence that the $n$\textsuperscript{th} sample corresponds to the groundtruth class $c^{(n)}$. 
In particular, the sample is correctly classified when $d^\BN(c^{(n)}, \vecx^{(n)}) > 1$.
Thus, the MM objective stimulates low classification error, while well calibrated class posteriors are deliberately not enforced.
In order to avoid that the model just optimizes the margins of a few samples, the sample-wise margin terms are capped by a hyper parameter $\gamma$, which is typically cross-tuned.    
 
An alternative and simple method for learning discriminative parameters are \emph{discriminative frequency estimates} \cite{Su2008DFE}.  According to this method, parameters are estimated using a perceptron-like algorithm, where parameters are updated by the prediction loss, i.e.\ the difference of the class posterior of the correct class (which is assumed to be $1$ for the data in the training set) and the class posterior according to the model using the current parameters. This type of parameter learning yields classification results comparable to those obtained by MCL~\cite{Su2008DFE}.
   
\item  \textbf{Hybrid Parameters.} Furthermore hybrid parameter learning which combines generative and discriminative objectives has been considered~\cite{Peh13}.
Hybrid parameters often achieve good prediction performance (due to the discriminative objective) while at the same time maintaining a generative character of the model, which is e.g.~beneficial under missing input features.
\end{itemize}

\subsubsection{Structure Learning}

The structure of a BN classifier, i.e.\ its graph $\graph$, encodes conditional independence assumptions.
Structure learning is naturally cast as a combinatorial optimization problem and is in general difficult --- even in the case where scores of structures decompose according to the network structure~\cite{chickering2004large}. 
For the generative case, several formal hardness results are available, e.g.\ learning polytrees~\cite{Dasgupta1999Polytrees} or learning general BNs~\cite{Heckerman1995LearningBayesianNetworks} are NP-hard optimization problems. 
Algorithms for learning generative structures often optimize some kind of penalized likelihood of the training data and try to determine the structure for example by performing independence tests~\cite{Friedman1997TANCMI}. Discriminative methods often employ local search heuristics~\cite{Pernkopf2010EfficientHeuristics, Pernkopf2011TANMM, Keogh99AugmentedBNC}. 

\begin{figure}[tb]
  \centering
  \begin{subfigure}[t]{0.24\textwidth}
    \centering
    \includegraphics[width=.99\textwidth]{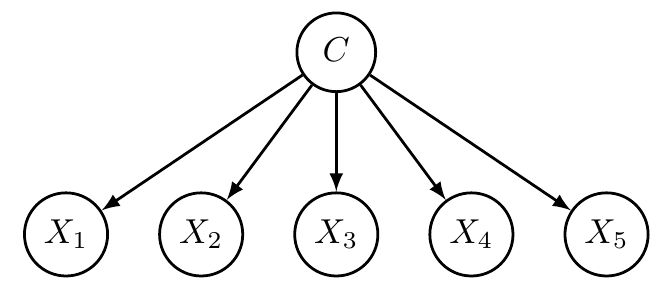}
    \caption{Naive Bayes structure}
    \label{fig:nb}
  \end{subfigure}
  \begin{subfigure}[t]{0.24\textwidth}
    \centering
    \includegraphics[width=.99\textwidth]{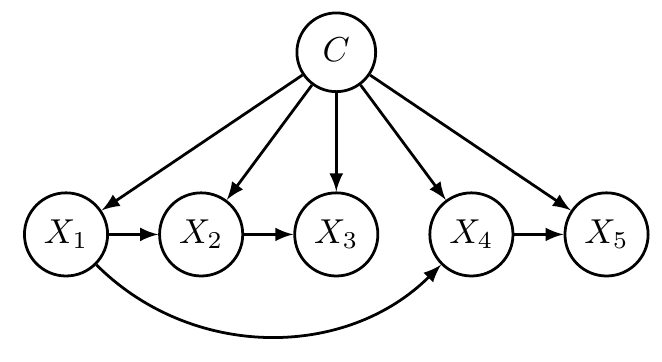}   
    \caption{TAN structure}
    \label{fig:tan}
  \end{subfigure}
  \caption{Exemplary BN structures used for classification.}
  \label{fig:examplaryBNstructures}
 \end{figure}

A good overview over different BN structures is provided in~\cite{Kol09}.
Here, we focus on relatively simple structures, i.e.\ the naive Bayes (NB) structure and tree augmented network (TAN) structures. 
 \begin{itemize}
  \item \emph{Naive Bayes (NB).} This structure implies conditional independence of the features, given the class, cf.~Figure~\ref{fig:nb}. Obviously, this conditional independence assumption is often violated in practice. Still, NB often yields impressively good performance in many applications~\cite{Zhang2004OptimalityNB}.
  \item \emph{Tree Augmented Networks (TAN).} This structure was introduced in~\cite{Friedman1997TANCMI} to relax the strong independence assumptions imposed by the NB structure and to enable better classification performance. In particular, each attribute may have at most one other attribute as an additional parent. An example of a TAN structure is shown in Figure~\ref{fig:tan}. TAN structures can be learned using generative and discriminative objectives~\cite{Pernkopf2010EfficientHeuristics, Pernkopf2011TANMM}.  
\end{itemize}
The reason for not using more expressive structures is mainly that more complex structures do not necessarily result in significantly better classification performance~\cite{Pernkopf2010EfficientHeuristics} while leading to models with many more parameters. 

\subsection{Reduced Precision in Bayesian Networks}   \label{sec:redPGMs}

Results from sensitivity analysis indicate that PGMs are well suited for low bit-widths implementations because they
are not sensitive to parameter deviations under the following two conditions~\cite{Chan02}.
Firstly, if the conditional probabilities are not too extreme, i.e. close to zero or one, and, secondly, if the posterior probabilities for different classes are significantly different.
Additionally, this is supported by empirical classification results for PGMs with reduced-precision parameters~\cite{Tsc14a, Tsc15, Tsc14} which we present in more detail in the following.

\begin{figure}
  \centering
  \begin{subfigure}[t]{0.22\textwidth}
    \centering
    \includegraphics[width=0.99\textwidth]{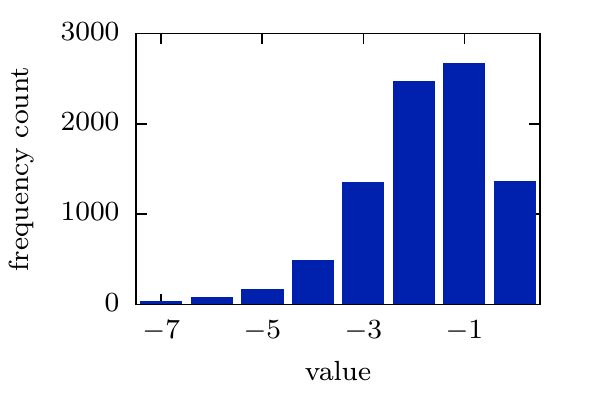}
    \caption{logarithmic probabilities}
    \label{fig:bn-red-prec-hist1}
  \end{subfigure}
  \begin{subfigure}[t]{0.22\textwidth}
    \centering
    \includegraphics[width=0.99\textwidth]{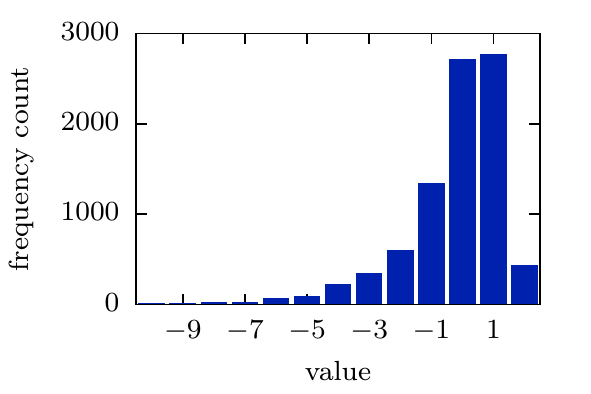}
    \caption{exponent of logarithmic conditional probabilities in double-precision}
    \label{fig:bn-red-prec-hist2}
  \end{subfigure}\\
  \begin{subfigure}[t]{0.22\textwidth}
    \centering
    \includegraphics[width=0.99\textwidth]{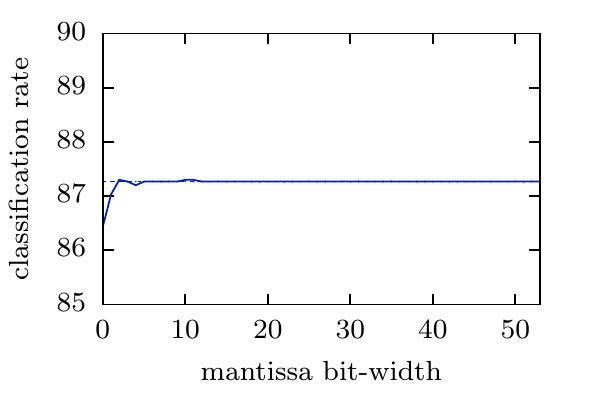}
    \caption{varying mantissa bit width, using full bit-width for exponent}
    \label{fig:bn-red-par-sweep1}
  \end{subfigure}
  \begin{subfigure}[t]{0.22\textwidth}
    \centering
    \includegraphics[width=0.99\textwidth]{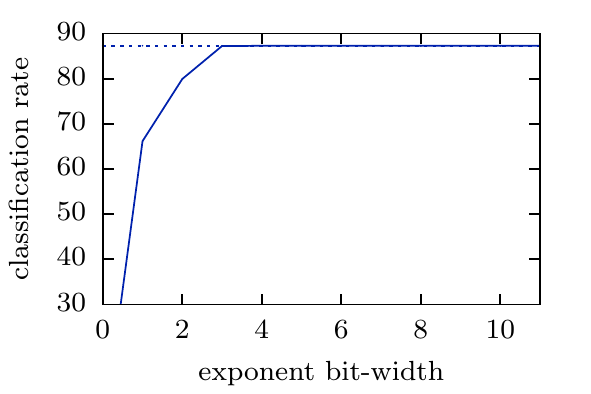}
    \caption{varying exponent bit width, using full bit width for mantissa}
    \label{fig:bn-red-par-sweep2}
  \end{subfigure}\\

  \caption{Top row: Histograms of (a) the log-parameters, and (b) the exponents of the log-parameters of a BN classifier for handwritten digit data with ML parameters assuming NB structure. Bottom row: Classification rates for varying bit widths of (a) the mantissa, and (b) the exponent, for handwritten digit data, NB structure, and log ML parameters. The classification rates using full double-precision logarithmic parameters are indicated by the horizontal dotted lines.}
  \label{fig:bn-req-prec}
\end{figure}

An important observation for the development of reduced-precision BNs is that for prediction it is sufficient to compute the product of conditional probabilities of the variables in the Markov blanket of the class variable $C$~\cite{Kol09}.
Equivalently, this corresponds to the computation of a sum of log-probabilities which can often be very efficiently implemented.
Furthermore, storing log-probabilities makes it easy to satisfy the condition from sensitivity analysis that it is important to be relatively accurate about small probabilities.
The required precision for storing reduced-precision floating point numbers depends on the range of values which the parameters of the BN assume, hence it is instructive to look at the histograms of log-probabilities for a BN trained for classifying handwritten digits.
Such histograms are shown in Figures~\ref{fig:bn-red-prec-hist1} and~\ref{fig:bn-red-prec-hist2}.
The log-probabilities assume only a small range of values, and considering the exponent of the corresponding (normalized) floating-point representation, we observe that there is only a small tail of large (negative) exponents, i.e.\ small probabilities.
This indicates, following the results of sensitivity analysis outlined above, that quantizing the log-probabilities should not reduce classification performance significantly.
Indeed, as illustrated in Figures~\ref{fig:bn-red-par-sweep1} and~\ref{fig:bn-red-par-sweep2} the performance of BNs with reduced-precision floating point numbers quickly reaches the performance of BNs with full-precision parameters with increasing parameter precision.

These illustrative results are for BNs with generatively optimized parameters, i.e.\ maximum-likelihood parameters.
This does not necessarily imply that BNs with discriminatively optimized parameters are also well-suited for reduced-precision parameters as discriminative parameters are in general more extreme, i.e.\ closer to zero or one.
However, Tschiatschek et al.~\cite{Tsc12} conducted an exhaustive evaluation of BNs with reduced-precision floating point parameters comparing BNs with generatively and discriminatively optimized parameters for the case in which these parameters are first estimated using full-precision floating point numbers and subsequently quantized to some desired (reduced) precision.
Their results indicate that BNs with discriminatively optimized parameters are almost as robust to precision reduction as BN classifiers with generatively optimized parameters.
Furthermore, even large precision reduction does not decrease classification performance significantly.
In general a mantissa with only 4 bits and a 5 bit exponent are sufficient to achieve close-to-optimal performance.
These findings are consistent among a large set of diverse data sets and BN structures.

\subsubsection{Learning Optimal Reduced-Precision Parameters}
\label{sec:pgm_opt_rp}

Reduced-precision BNs for classification achieve remarkable performance when these parameters are obtained by rounding full-precision parameters.
Nevertheless, a natural question that arises is whether improved performance can be achieved by learning parameters that are tailored for reduced precision.
This question was affirmatively studied in~\cite{tschiatschek2014diss,Tsc15}.
The authors proposed a branch-and-bound algorithm for finding globally optimal discriminative fixed-precision parameters.
The resulting parameters have superior classification performance compared to parameters obtained by simple rounding of double-precision parameters, particularly for very low number of bits, cf.\ Section~\ref{sec:mnist}.
Again, these findings are consistent among a large set of diverse data sets and BN structures~\cite{tschiatschek2014diss,Tsc15}.

\subsubsection{Online Learning in Reduced Precision}

While in many applications suitable reduced-precision parameters for BNs can be precomputed using the techniques outlined in the previous section, there are applications requiring to learn parameters \emph{within} the application, i.e.~on a system supporting only reduced-precision computations.
Examples include applications requiring fine-tuning of parameters for domain adaptation or adaptation of parameters to user preferences.
Thus it is important to enable learning reduced-precision parameters for BNs using reduced-precision computations only.
In \cite{tschiatschek2014diss}, this setting was investigated.
The authors propose algorithms for learning ML parameters and for learning MM parameters.

The algorithms are developed for the \emph{online setting}, i.e.\ when parameters are updated on a per-sample basis. 
In this setting, learning using reduced-precision computations requires specialized algorithms because gradient-descent (or gradient-ascent) procedures using reduced-precision arithmetic typically do not perform well.
The problem is resolved by using precomputed lookup tables of small sizes for log-parameters which can be efficiently indexed by keeping and (on overflows) scaling feature counts.
The resulting algorithms have very low computational demands, mainly requiring counters and a little memory for storing the lookup tables.
At the same time the proposed algorithms yield parameters with close-to-optimal performance while only having slightly slower convergence than comparable algorithms using full-precision arithmetic.

\section{Experimental Results}   \label{sec:experiments}

In this section, we first exemplify the trade-off between model performance, memory footprint and computation time on the CIFAR-10 classification task in Section~\ref{sec:cifar}.
This example highlights that finding a suitable balance between these requirements remains challenging due to diverse hardware and implementation issues.
Furthermore, we provide an extensive comparison between a rich collection of hardware-efficient approaches discussed in this paper, on the challenging task of ImageNet classification in Section~\ref{sec:imagenet}.
In Section~\ref{sec:speech} we present a real-world speech enhancement example, where hardware-efficient BNNs have led to dramatic memory and computation time reductions. Section~\ref{sec:mnist} shows exemplary results comparing PGMs and DNNs on the classical MNIST data set. 
The focus here is on prediction performance and the number of bits necessary to represent the models. We conclude the experimental section with an example of randomly missing features during model testing (see Section~\ref{sec:uncertainty}). Such scenarios can be easily treated with probabilistic models.

\subsection{Prediction Accuracy, Memory Footprint and Computation Time Trade-Off}
\label{sec:cifar}
To exemplify the trade-off to be made between memory footprint, computation time and prediction accuracy, we implemented general matrix multiply (GEMM) with variable-length fixed-point representation on a mobile CPU (ARM Cortex A15), exploiting its NEON SIMD instructions.
Using this implementation, we ran a 32-layer ResNet NN with custom quantization on weights and activations representation \cite{Zhou2016} and compare these results with single-precision floating point.
We use the CIFAR-10 data set, containing color images of 10 object classes (airplanes, automobiles, birds, cats, deer, dogs, frogs, horses, ships and trucks).

\begin{figure}[t]
\begin{center}
 \includegraphics[width=8cm]{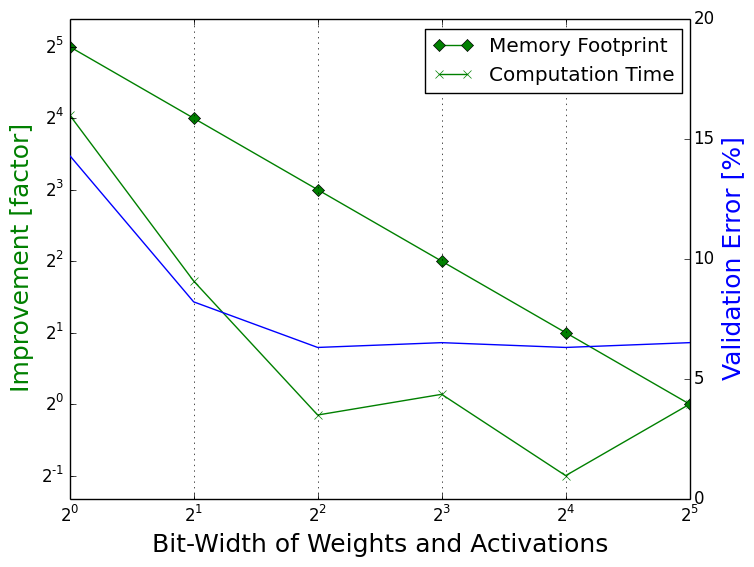}
\caption{Improvement of reduced precision over single-precision floating point on memory footprint and computation time (green) and the respective validation error of ResNet-32 on CIFAR-10 (blue).}
 \label{Tradeoff}
\end{center}
\end{figure}

Figure~\ref{Tradeoff} reports the impact of reduced precision on runtime, memory requirements and classification accuracy, averaged over the test set.
As can be seen, reducing the bit width to 16, 8 or 4 bits does not improve runtimes and even is harmful in the case of 4 bits.
The reason for this behavior is that our implementation uses bit-width doubling for these precisions in order to ensure correct GEMM computations.
Since bit widths of 2 and 1 do not require bit-width doubling, we obtain runtimes close to the theoretical linear speed-ups.
In terms of memory footprint, our implementation evidently reaches the theoretical linear improvement.
While reducing the bit width of weights and activations to only 1 or 2 bits improves memory footprint and computation time significantly, these settings also show decreased performance.
In this example, the sweet spot appears to be 2 bit precision, but also the predictive performance for 1 bit precision might be acceptable for some applications.
This extreme setting is evidently beneficial for highly constrained scenarios and is easily exploited on today's hardware, as shown in the following section.

\subsubsection{Computation Savings for BNNs}   \label{sec:computational_savings}

In order to show that the advantages of binary computation translate to other general-purpose processors, we implemented matrix-multiplication operators for NVIDIA GPUs and ARM CPUs. 
Classification in BNNs can be implemented very efficiently as 1-bit scalar products, i.e.~multiplications of two vectors $\mathbf{x}$ and $\mathbf{y}$ of length $N$ reduce to bit-wise \textit{xnor()} operation, followed by counting the number of set bits with \textit{popc()}:
\begin{equation}\label{eq:xnor}
\mathbf{x}\cdot \mathbf{y} = N-2*popc(xnor(\mathbf{x},\mathbf{y})), x_i, y_i \in{[-1, +1] }\quad\forall{i}.
\end{equation}
We use the matrix-multiplication algorithms of the MAGMA and Eigen libraries and replace float multiplications by \textit{xnor()} operations, as depicted in Equation~\eqref{eq:xnor}. 
Our CPU implementation uses NEON vectorization in order to fully exploit SIMD instructions on ARM processors. 
We report execution time of GPUs and ARM CPUs in Table \ref{tab:performance_metrics}.  
As can be seen, binary arithmetic offers considerable speed-ups over single-precision with manageable implementation effort. 
This also affects energy consumption since binary values require less off-chip accesses and operations. Performance results of x86 architectures are not reported because neither SSE nor AVX ISA extensions support vectorized \textit{popc()}. 

\begin{table}[htb]
  \centering
	\caption{Performance metrics for matrix $\cdot$ matrix multiplications on a NVIDIA Tesla K80 and ARM Cortex-A57.}
    \label{tab:performance_metrics} 
    \begin{tabular}{|l|l l l l|}
      \hline
      arch  & matrix size & time (float32) & time (binary) & \bf speed-up\\
      \hline
	  GPU & 256 & 0.14ms & 0.05ms & \bf2.8\\	
	  GPU & 513 & 0.34ms & 0.06ms & \bf5.7\\
	  GPU & 1024 & 1.71ms & 0.16ms & \bf10.7\\	
	  GPU & 2048 & 12.87ms & 1.01ms & \bf12.7\\	
      \hline
	  ARM & 256 & 3.65ms & 0.42ms & \bf8.7\\	
	  ARM & 513 & 16.73ms & 1.43ms & \bf11.7\\
	  ARM & 1024 & 108.94ms & 8.13ms & \bf13.4\\
	  ARM & 2048 & 771.33ms & 58.81ms & \bf13.1\\
      \hline	
    \end{tabular}
\end{table}

While improvements of memory footprint and computation time are independent of the underlying tasks, the prediction accuracy highly depends on the complexity of the data set and the used neural network.
Simple data sets such as MNIST, allow for aggressive quantization without affecting prediction performance significantly, while binary/ternary quantization results in  severe prediction degradation on more complex data sets, such as ImageNet.

\subsection{Resource-Efficient DNNs on ImageNet}
\label{sec:imagenet}
We compare the performance of different quantization strategies on the example of AlexNet \cite{Krizhevsky2012} on the ImageNet ILSVRC-2012 data set \cite{ImageNet09}.
Since 2010, ImageNet is the data set for the annual competition called the Large-Scale Visual Recognition Challenge (ILSVRC).
ILSVRC uses a subset of ImageNet with roughly 1000 images in each of 1000 categories, comprising roughly 1.2M training and 50k validation images with high resolution.
The ILSVRC is considered to be one of the most challenging data sets for DNNs and, consequently, for quantization.
It is common practice to report two prediction-accuracy rates: Top-1 and Top-5 accuracy, where Top-5 is the fraction of test images for which the correct label is among the five labes.
Table \ref{alexnet} reports the accuracy gap (Top-1 and Top-5) between single-precision floating point and the respective quantization approach.

\begin{table*}[]
\centering
\caption{Accuracy gap between single-precision floating point and different state-of-the-art quantization approaches of AlexNet on ImageNet for different bit-width combinations of Activations (A) and Weights (W).}
\label{alexnet}
\begin{tabular}{|c|c|c|c|c|c|c|c|c|c|c|c|c|c|}
					\hline
                   A-W & Gap & DC & BC & BNN & XNOR & DoReFa & TWN & TTQ & QNN & HWGQ & SYQ & TSQ & DeepChip \\ \hline
					\multirow{2}{*}{32-32}
					& Top-1  & 57.2 & 56.6 & 56.6 & 56.6 & 57.2 & 57.2  & 57.2 & 56.6 & 58.5 & 56.6 & 58.5 & 56.2 \\\cline{2-14}
					& Top-5  & 80.3 & 80.3 & 80.2 & 80.2 & 80.3 & 80.3 & 80.3 & 80.2 & 81.5 & 80.2 & 81.5 & 78.3 \\\hline
					\multirow{2}{*}{32-8/5}
					& Top-1  & 0.0 & -- & -- & -- & -- & -- & -- & -- & -- & -- & -- & -- \\\cline{2-14}
					& Top-5  & 0.0 & -- & -- & -- & -- & -- & -- & -- & -- & -- & -- & -- \\\hline
					\multirow{2}{*}{32-2}
					& Top-1  & -- & -- & -- & -- & -- & -2.7 & +0.3 & -- & -- & -- & -- & -- \\\cline{2-14}
					& Top-5  & -- & -- &-- & -- & -- & -3.5 & -0.6 & -- & -- & -- & -- & -- \\\hline
					\multirow{2}{*}{32-1}
					& Top-1  & -- & -21.2 & -- & +0.2 & -- & -- & -- & -- & -- & -- & -- & -- \\\cline{2-14}
					& Top-5  & -- & -19.3 & -- & -0.8 & -- & -- & -- & -- & -- & -- & -- & -- \\\hline
					\multirow{2}{*}{8-2}
					& Top-1  & -- & -- & -- & -- & -- & -- & -- & -- & --  & +1.5 & -- & +0.2 \\\cline{2-14}
					& Top-5  & -- & -- & -- & -- & -- & -- & -- & -- & -- & +0.6 & -- & +0.7 \\\hline
					\multirow{2}{*}{2-2}
					& Top-1  & -- & -- & -- & -- & -- & -- & -- & -- & --  & -0.8 & -0.5 & -- \\\cline{2-14}
					& Top-5  & -- & -- & -- & -- & -- & -- & -- & -- & -- & -1.0 & -1.0 & -- \\\hline
					\multirow{2}{*}{2-1}
					& Top-1  & -- & -- & -- & -- & -- & -- & -- & -5.6 & -5.8 & -1.2 & -- & -- \\\cline{2-14}
					& Top-5  & -- & -- & -- & -- & -- & -- & -- & -6.5 & -5.2 & -1.6 & -- & -- \\\hline
					\multirow{2}{*}{1-1}
					& Top-1  & -- & -- & -28.7 & -12.4 & -11.8 & -- & -- & -- & -- & -- & -- & --  \\\cline{2-14}
					& Top-5  & -- & -- & -29.8 & -11.0 & -11.0 & -- & -- & -- & -- & -- & -- & -- \\\cline{1-14}
\end{tabular}
\end{table*}

First, we compare several strategies that quantize weights of a DNN on the basis of the Top-1 accuracy gap.
Deep compression (DC) \cite{Han2016} effectively reduces weights (using weight sharing) to 8 bit (convolutional layers) and 5 bit (fully-connected layers) in order to obtain full-precision prediction accuracy.
Binarization of weights was first introduced by binary connect (BC) \cite{Courbariaux2015b} with about a -21.2\% Top-1 accuracy gap.
The introduction of scaling coefficients by XNOR-Net \cite{Rastegari2016} outperformed BC by a large margin with prediction performance close to full-precision weights.
Quantizing weights to a ternary representation is superior to a binary representation on large-scale data sets.
TWN \cite{Li2016} reduced the gap to full precision AlexNet to only -2.7\% and TTQ \cite{Zhu2017} even outperformed full-precision by 0.3\%.
SYQ \cite{Faraone18} further improves ternary quantization by using pixel-wise instead of layer-wise scaling coefficients.
At the cost at a higher memory footprint, they are able to outperform Top-1 and Top-5 prediction performance of single-precision floating point by 1.5\% and 0.6\% respectively.

Whereas binarization of weights works well on AlexNet, binarization of activations shows severe performance degradation (-28.7\% and -12.4\% Top-1 accuracy gap for BNN \cite{Hubara2016} and XNOR-Net \cite{Rastegari2016}, respectively).
QNNs \cite{Hubara2016b} and HWGQ \cite{Cai2017} tackle this problem by using more bits for activations while binarizing weights: for instance, using 2-bit activations decreases the Top-1 gap to -5.6\% for QNN and -5.8\% for HWGQ.
TSQ \cite{Wang18} further improves the approach of HWGQ and achieves -0.5\% Top-1 gap (with ternary weights and 2-bit activations).
SYQ and DeepChip \cite{DeepChip18} require 8-bit fixed point in order to maintain full-precision accuracy.

\subsection{A Real-World Example: Speech Mask Estimation using Reduced-Precision DNNs}   \label{sec:speech_mask}
\label{sec:speech}
\begin{figure}[!ht]
\centering
\includegraphics[width=0.5\textwidth]{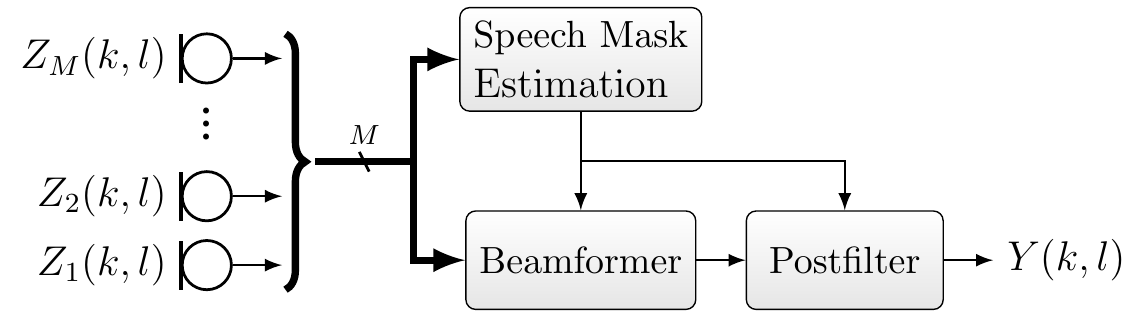}
\caption{System overview, showing the microphone signals $Z_m(k,l)$ and the beamformer+postfilter output $Y(k,l)$ in frequency domain.}
\label{fig:system_overview}
\end{figure}

We provide a complete example employing hardware-efficient BNNs applied to acoustic beamforming, an important component for various speech enhancement systems.
A particularly successful approach employs DNNs to estimate a speech mask, i.e.~a speech presence probability of each time-frequency cell.
This speech mask is used to determine the power spectral density (PSD) matrices of the multi-channel speech and noise signals, which are subsequently used to obtain a beamforming filter such as the minimum variance distortionless response (MVDR) beamformer or generalized Eigenvector (GEV) beamformer \cite{Warsitz:Jul07, Warsitz:08, Heymann:2016,Heymann:dec15, Erdogan:sep16, Pfei:Mar17}.
An overview of a multi-channel speech enhancement setup is shown in Figure~\ref{fig:system_overview}.

In this experiment, we compare single-precision DNNs and BNNs trained with STE \cite{Hubara2016} for the estimation of the speech mask. 
For both architectures, the dominant Eigenvector of the noisy speech PSD matrix~\cite{Pfei:Mar17} is used as feature vector, where it is quantized to 8 bit integer values for the BNN. 
As output layer, a linear activation function is used, which reduces to counting the binary neuron outputs, followed by normalization to yield the speech presence probability mask $\text{p}_{SPP} \in [0,1]$. 
Further details of the experimental setting can be found in~\cite{Zoh18}.

\subsubsection{Data and Experimental Setup}   \label{ssec:database}

For evaluation we used the CHiME corpus \cite{chime3overview} which provides 2 and 6-channel recordings of a close-talking speaker corrupted by four different types of ambient noise. 
Ground truth utterances (i.e. the separated speech and noise signals) are available for all recordings, such that the ground truth speech masks $\text{p}_{SPP,opt}(k, l)$ at time $l$ and frequency bin $k$ can be computed. 
In the test phase, the DNN is used to predict $\hat{\text{p}}_{SPP}(k, l)$ for each utterance, used to estimate the corresponding beam-former. 
A single-precision 3-layer DNN with 513 neurons per layer and BNNs with 513 and 1024 neurons per layer are used. 
The DNNs were trained using ADAM \cite{Kingma:Jul15} with default parameters and a dropout probability of 0.25.

\subsubsection{Speech Mask Accuracy}  \label{sssec:mask_accuracy}

\begin{figure}[]
\centering
\includegraphics[width=0.5\textwidth]{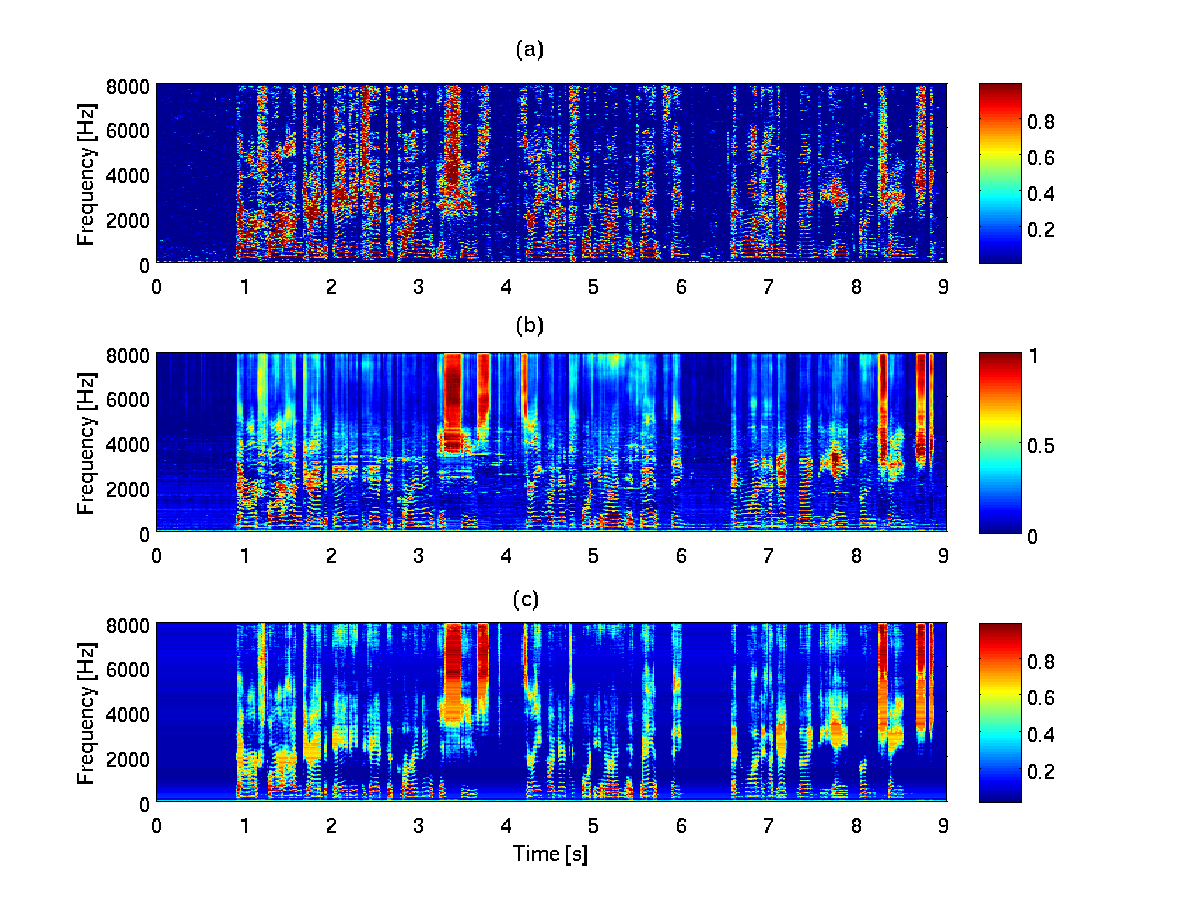}
\caption{Speech presence probability mask: (a) Optimal speech mask ${\text{p}}_{SPP,opt}(k,l)$; (b)  Prediction of $\hat{\text{p}}_{SPP}(k,l)$ using DNNs with 513 neurons/layer; (c) Prediction of $\hat{\text{p}}_{SPP}(k,l)$ using BNNs with 1024 neurons/layer.}
\label{fig:speech_mask}
\end{figure}

Figure \ref{fig:speech_mask} shows the optimal and predicted speech masks of the DNN and BNN for an example utterance (F01\_22HC010W\_BUS). 
We see that both methods yield very similar results and are in good agreement with the ground truth.
Table \ref{tab:mse} reports the prediction error 
$\mathcal{L} = \frac{100}{KL} \sum_{k=1}^K \sum_{l=1}^L \big| \hat{\text{p}}_{SPP}(k,l) - \text{p}_{SPP,opt}(k,l) \big|$ in [\%].
Although single-precision DNNs achieved the best prediction error on the test set, they do so only by a small margin.
Doubling the networks size of BNNs slightly improved the error on the test set for the case of 6 channels.

\begin{table}[htb]
	\caption{Mask prediction error $\mathcal{L}$ in [\%] for DNNs with 513 neurons/layer and BNNs with 513 or 1024 neurons/layer.}
    \label{tab:mse}
  \centering
    \begin{tabular}{|l|l|l| r r r|}
      \hline
      model & neurons / layer & channels & \bf train & \bf valid & \bf test \\	    
      \hline
	  DNN & 513 & 2ch & 5.8 & 6.2 & 7.7 \\
	  BNN & 513 & 2ch &  6.2 & 6.2 & 7.9 \\
	  BNN & 1024 & 2ch & 6.2 & 6.6 & 7.9 \\	  
      \hline
	  DNN & 513 & 6ch & 4.5 & 3.9 &  4.0 \\
	  BNN & 513 & 6ch & 4.7 & 4.1 & 4.4 \\	  
	  BNN & 1024 & 6ch & 4.9 & 4.2 & 4.1 \\	  
      \hline
    \end{tabular}
\end{table}

\subsubsection{Perceptual Audio Quality}
\label{sssec:audio_quality}

Given the predicted speech mask $\hat{\text{p}}_{SPP}(k,l)$, we construct the GEV-PAN beamformer \cite{Pfei:Aug17} for both the 2 and 6-channel data. The overall perceptual score (OPS) \cite{Emiya:Sep11} is used to evaluate the performance of the resulting speech signal $Y(k,l)$ in terms of perceptual speech quality. Ground truth estimates required for these scores are obtained using the $\text{p}_{SPP,opt}(k,l)$ and the GEV-PAN.

Table \ref{tab:OPS_scores} reports the OPS given the enhanced utterances of the GEV-PAN beamformer. 
GEV-PAN outperforms the CHiME4-baseline enhancement system, i.e. the BeamformIt!-toolkit \cite{chime3overview}, and the front-end of the best CHiME3 system~\cite{Higuchi:Mar16}, i.e.\ CGMM-EM. Doubling the network size of BNNs mostly improves the OPS scores. In 
general, BNNs achieve on average only a slightly lower OPS score than the single-precision DNN baseline. 

\begin{table}[htb]
	\caption{Overall perceptual score (OPS) for various beamformers (BeamformIt, GEV-PAN, MVDR) using DNNs and BNNs for speech mask estimation.}
    \label{tab:OPS_scores}
\begin{scriptsize}
  \resizebox{8.5cm}{!}{
    \begin{tabular}{|l|l r r r|}
      \hline
      method & set & \bf train & \bf valid & \bf test \\
      \hline
	  CHiME4 baseline  & simu & 33.11 & 34.73 & 31.46 \\
	  (BeamformIt), 5ch \cite{chime3overview}  & real & 29.97 & 36.45 & 36.74 \\
      \hline
	  CGMM-EM with MVDR     & simu & 52.15 & 43.02 & 40.59 \\
	  and postfilter, 6ch \cite{Higuchi:Mar16}        & real & 44.95 & 41.89 & 36.87 \\
      \hline	  
      \hline
	  \bf DNN (513 neurons / layer)  & simu & \bf 64.21 & \bf 61.74 & \bf 56.32 \\
	  \bf with GEV-PAN, 2ch      & real & \bf 64.21 & \bf 62.72 & \bf 56.32 \\
      \hline
	   BNN (513 neurons / layer)   & simu &  58.11 &  57.58 &  57.58 \\
	   with GEV-PAN, 2ch     & real &  56.79 &  57.52 &  41.24 \\	
      \hline
	  \bf BNN (1024 neurons / layer)   & simu & \bf 61.64 & \bf 60.78 & \bf 54.20 \\
	  \bf with GEV-PAN, 2ch     & real & \bf 61.64 & \bf 60.78 & \bf 45.22 \\	
      \hline
	  \bf DNN (513 neurons / layer) ,  & simu & \bf 67.98 & \bf 66.76 & \bf 68.71 \\
	  \bf with GEV-PAN 6ch      & real & \bf 69.98 & \bf 70.33 & \bf 63.28 \\
      \hline
	   BNN (513 neurons / layer)  with  & simu &  61.44 &   55.87 &  62.39 \\
	   with GEV-PAN, 6ch      & real &  63.03 &  64.77 &  64.52 \\      	    
      \hline
      	  \bf BNN (1024 neurons / layer)   & simu & \bf 65.59 & \bf  64.98 & \bf 68.41 \\
	  \bf with GEV-PAN, 6ch      & real & \bf 67.91 & \bf 68.41 & \bf 59.94 \\      	    
      \hline
    \end{tabular}
  }
    \end{scriptsize}
\end{table}

\subsection{Resource-efficient DNNs and PGMs on MNIST}
\label{sec:mnist} 
While PGMs with sparse structures such as NB or TAN are usually computationally efficient and have a small memory footprint, they often do not achieve the same prediction performance as DNNs.
However, PGMs have advantages in important settings for applying machine learning in ``the wild'', e.g.\ when a considerable number of input features is missing.
Both modeling approaches have rich capabilities for machine learning on embedded devices.
Here, the classification performance of both reduced-precision PGMs and DNNs is compared on the MNIST data.

\subsubsection{Data}
\begin{figure}[!ht]
\centering
\includegraphics[width=0.45\textwidth]{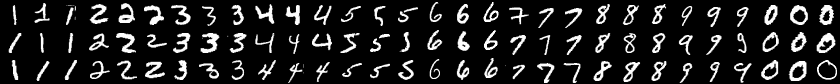}
\caption{Samples of MNIST.}
\label{fig:mnist_samples}
\end{figure}

The MNIST data set for handwritten digit recognition \cite{LeCun1998} contains 60000 training images and 10000 test images of size $28 \times 28$ with gray-scale values. Some samples from the data set are shown in Figure \ref{fig:mnist_samples}.
For DNNs the training set is further split into 50000 training samples and 10000 validation samples. Each pixel is treated as feature, i.e.\ $\mathbf{x} \in \mathbb{R}^{784}$.
For PGMs and \emph{small-size} DNNs the data is down-sampled by a factor of two, resulting in a resolution of $14 \times 14$ pixels, i.e.\ $\mathbf{x} \in \mathbb{R}^{196}$.

\subsubsection{Results}
We report the performance of reduced-precision DNNs with sign activations using variational inference (NN VI) \cite{Roth2018} and using the STE (NN STE) \cite{Hubara2016}. As a baseline, we compare with real-valued (32 bit) DNNs (NN real) trained with batch normalization \cite{Ioffe2015}, dropout \cite{Srivastava2014}, and ReLU activations. A three layer structure with $1200-1200$ hidden units is selected. Several hyperparameters for all methods were tuned using 50 iterations of Bayesian optimization \cite{Snoek2012} on a separate held-out validation set.
For NN VI, we used 3-bit weights for the input layer and ternary weights $w \in \{-1,0,1\}$ for the remaining layers. During training, dropout was used to regularize the model. Results for the most probable model from the approximate posterior $\mathbf{W} = \argmax_{\mathbf{W}} q(\mathbf{W})$ are reported.
For NN STE, the weights in the input layer were quantized to 3-bit weights as above, and the weights of the remaining layers were always quantized to binary values. In addition to dropout, NN STE also uses batch normalization which appears to be a crucial component here. Although batch normalization requires real-valued parameters, it merely results in a shift of the sign activation function that introduces only a marginal computational overhead at test time \cite{Umuroglu2017}.

We contrast the results for the reduced-precision DNNs with those of BNs. In particular, the NB structure and MM parameter learning has been used (see Section~\ref{sec:pgms}).

The classification errors (CE) [\%], the model size (\#Param) [kbits] and the model configuration for the DNNs and BN are shown in Table \ref{tab:nnvidiscrete_errors}.
NN STE (3-bit) performs on par with NN VI (3-bit) while NN (real) with a three layer structure (i.e. 1200-1200 neurons) slightly outperforms both.
The classification performance of the BN is worse compared to DNNs. However, the achieved performance is impressive considering that only $6{,}720$ parameters (each represented by 6-bits) are used\footnote{After discretizing input features and the removal of features with constant values across the data set.}, i.e.\ the BN is a factor of ~60, ~120, ~1900 smaller than NN STE, NN VI and NN real using a 3 layer structure with $1200-1200$ hidden units, respectively.

Moreover, we scaled the NN STE down to about the same model size of 40 kbits as the BN with NB structure and MM parameters using the down-sampled MNIST data (BN NB MM).
Results show that NN STE slightly outperforms the BN when using batch normalization. Furthermore, NN STEs have one hidden layer, while the BN is \emph{shallow} which might explain some of the performance gain. Better performance with BNs can be achieved with more expressive structures, e.g.\ a BN classifier with TAN-MM structure has $31{,}399$ parameters and achieves a classification error of about $4{,}75\%$~\cite{Pernkopf2011TANMM}.
Furthermore, classification in BNs is computationally extremely simple -- just the joint probability $\pdfBN(C,\mathbf{X})$ has to be computed. This amounts to summing up the log conditional probabilities for each feature $X_i$. These results suggest that BNs enable a good trade-off between computational requirements for inference, memory demands and prediction performance. Additionally, they are advantageous in case of missing input features. This is shown in Section~\ref{sec:uncertainty}.
 
\begin{table}[ht!]
\scriptsize
\centering
\caption{Classification errors (CE) [\%], model size (\#Param) [kbits] and model configuration for different NN models and BNs. NN real: Real-valued DNNs;  NN STE: DNNs with 1 or 3 bit weights in the first layer, binary weights in the remaining layers. NN VI: DNNs with 1 or 3 bit weights in the first layer, ternary weights in the remaining layers. BN: Bayesian network with naive Bayes (NB) structure and MM parameters using 6-bit parameters.}
\begin{tabular}{|l| c|c |c |c |c |c|}
 \hline 
 Classifier & CE  & (\#Param) & input & input & batch & layers/  \\ 
	    &	[\%]	& [kbits]	& size & layer & norm & neurons \\ \hline
 NN real          &   0.87 & ~76 800  & $28 \times 28$ & 32-bit & yes & 1200-1200 \\ \hline
 NN STE      &   1.24 & ~ 2 550 & $28 \times 28$ & 3-bit & yes & 1200-1200\\ 
 NN VI        &   1.28 & ~ 4 790 & $28 \times 28$ & 3-bit & no & 1200-1200\\ \hline \hline
 BN NB MM    &	6.72 & ~ 40  & $14 \times 14$ & - & - & -\\ \hline \hline
 NN STE     & 4.25 &  ~ 40 & $14 \times 14$ & 3-bit & yes & 65 \\
 NN STE     & 7.82 &  ~ 40 & $14 \times 14$ & 3-bit & no & 65 \\
 NN STE     & 3.72 &  ~ 40 & $14 \times 14$ & 1-bit & yes & 193 \\
 NN STE     & 6.99 &  ~ 40 & $14 \times 14$ & 1-bit & no & 193 \\ \hline
\end{tabular}
\label{tab:nnvidiscrete_errors}
\end{table}

\begin{figure}
  \centering
  \begin{subfigure}[t]{0.45\textwidth}
    \centering
    \includegraphics[width=0.80\textwidth]{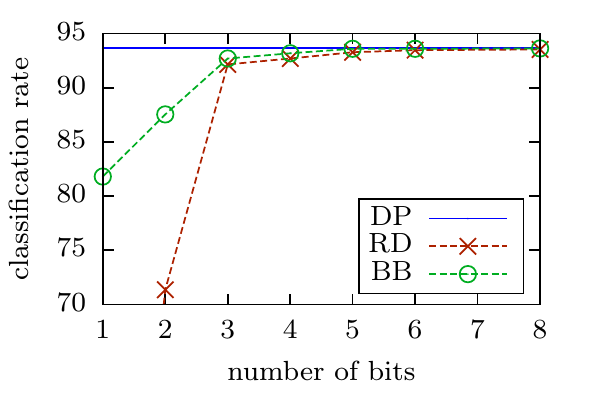}
\end{subfigure}

  \caption{Classification rates of BN classifiers using NB structure for MNIST and discriminative MM parameters. Parameters computed using the branch-and-bound approach (BB)~\cite{tschiatschek2014diss} outperform parameters obtained by first computing full-precision parameters and subsequently rounding them to the desired precision (RD). The classification rates using double precision parameters (DP) upper bounds the performance of the classifiers with reduced-precision parameters.}
  \label{fig:bn-bb}
\end{figure}

The classification results for BN NB MM over various bit width is shown in Figure~\ref{fig:bn-bb}. In particular, we show results for full-precision floating point parameters, reduced-precision fixed-point parameters obtained by rounding and optimal reduced-precision fixed-point parameters obtained by the algorithm outlined in Section~\ref{sec:pgm_opt_rp}. The performance of the reduced-precision parameters quickly approaches that of full-precision parameters with an increasing number of bits. The optimal reduced-precision parameters achieve improved performance over the reduced-precision parameters obtained by rounding, in particular for low numbers of bits.

\subsection{Uncertainty Treatment}
\label{sec:uncertainty}

A key advantage of probabilistic models is that they allow to treat uncertainty in a consistent manner.
While there are many types of uncertainty \cite{Ghahramani2015}, e.g.~data uncertainty stemming from noise, predictive uncertainty stemming from ambiguities, or model uncertainty, all of these can be treated in a uniform manner by virtue of probabilistic inference.

As an example, consider that a classifier has been trained on a fully observed data set (i.e.~there are no input features missing), but it shall be applied in a setting where inputs drop out at random. 
This missing-at-random (MAR) scenario \cite{Little2014}, although being an arguably simple and common one, is still a major cause of trouble for purely discriminative approaches like DNNs.
The problem here is that DNNs at best represent a conditional distribution $p(C \,|\, \mathbf{X})$, which does not capture any correlations within $\mathbf{X}$.
In a full joint distribution $p(C, \mathbf{X})$, as represented by PGMs, the MAR scenario is naturally handled by marginalizing missing features.
In particular, given values $\mathbf{x}_o$ for a subset of input features $\mathbf{X}_o \subset \mathbf{X}$, we use $p(C \,|\, \mathbf{x}_o) = \frac{\int p(C, \mathbf{x}_h, \mathbf{x}_o) d\mathbf{x}_h}{\sum_c \int p(c, \mathbf{x}_h, \mathbf{x}_o) d\mathbf{x}_h}$ for classification, where $\mathbf{X}_h = \mathbf{X} \setminus \mathbf{X}_o$.

There is, however, a hinge for PGMs trained in a discriminative way: while discriminative training generally improves classification results on completely observed data, we cannot expect that these models also are robust under missing inputs.
This can be easily seen by factorizing the joint $p(C, \mathbf{X})$ into $p(C \,|\, \mathbf{X})p(\mathbf{X})$.
Discriminative learning deliberately ignores $p(\mathbf{X})$ and focuses on tuning $p(C \,|\, \mathbf{X})$.
In order to treat missing inputs in a consistent manner, however, we need to faithfully capture $p(\mathbf{X})$ as well.
To this end, we might use hybrid generative-discriminative methods \cite{Peh13,ROTH2018Hybrid}, which aim at a sensible trade-off between predictive accuracy and ``generativeness'' of the employed model.

\begin{figure}
  \centering
  \includegraphics[width=0.45\textwidth]{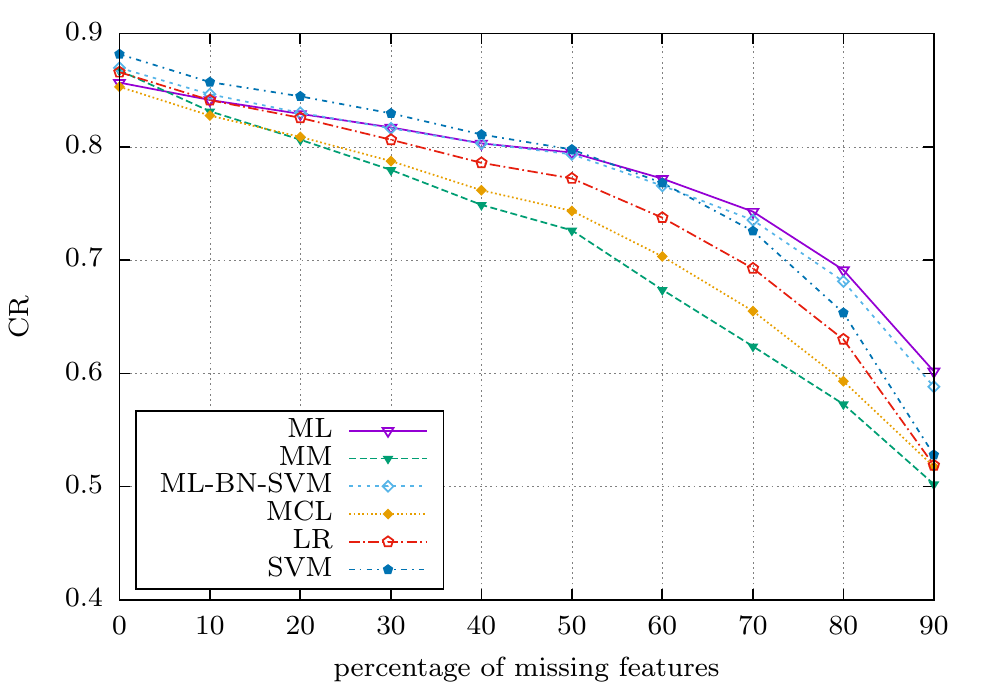}
  \caption{Classification rate (CR) of various models under missing input features. 
  ML, MM, ML-BN-SVM, and MCL are BNs trained with maximum-likelihood, maximum margin \cite{Per11}, a hybrid objective \cite{Peh13} and maximum-conditional likelihood, respectively.
  LR is logistic regression and SVM is a kernelized support vector machine. 
  For LR and SVM, missing features were treated using 5-nearest neighbor imputation; for BNs missing features were treated by marginalization.}
  \label{fig:missing_features}
\end{figure}

We demonstrate this effect using BNs trained with ML, MM, a hybrid ML-MM objective (ML-BN-SVM) \cite{Peh13}, and MCL, and compare with classical logistic regression (LR) and kernelized SVMs.
Since LR and SVMs cannot deal with missing inputs, we used 5-nearest neighbor imputation before applying the model.
In Figure~\ref{fig:missing_features} we show the classification rate of all models as a function of percentage of missing features, averaged over 25 UCI data sets.
We see that the model trained with ML is most robust under missing input features, and for more than 60\% of missing features it outperforms all other models.
The hybrid solution ML-BN-SVM has the second highest accuracy under no missing features and is almost as robust as the purely generative solution.
The purely discriminative models MM, MCL, LR, and SVM are clearly more sensitive to missing features.
Furthermore, note that K-NN imputation for LR and SVM requires that the training set is available also during test time; thus, this approach to treat missing features also comes with an significant additional memory requirement.

\section{Conclusion}
We compared deep neural networks (DNNs) and probabilistic graphical models (PGMs) regarding their efficiency and robustness for real-world systems, focusing on the possible trade offs of computation/memory demands and prediction performance.
In general, DNNs require large amounts of computational and memory resources while
PGMs with sparse structure are usually computationally efficient and have a small memory footprint.
Unfortunately, PGMs often do not accomplish the same prediction performance as DNNs do, but they are able to treat uncertainty in a natural way and show benefits in case of missing features. Both modeling approaches have rich capabilities for machine learning on embedded devices. 

For DNNs we discussed approaches for model size reduction. Furthermore, a comprehensive overview of DNNs with reduced-precision parameters was provided, with a focus on binary and ternary weights. 

For PGMs we summarized discriminative and hybrid parameter and structure learning techniques to improve the prediction performance. Furthermore we devoted a section on PGMs using reduced-precision parameters. 
In experiments, we demonstrated the trade-off between prediction performance and computational- and memory requirements for several challenging machine learning benchmark data sets. Furthermore, we presented exemplary results comparing reduced-precision PGMs and DNNs.

\section*{Acknowledgments}
This work was supported by the Austrian Science Fund (FWF) under the project number I2706-N31 and the German Research Foundation (DFG). 
Furthermore, we acknowledge the LEAD Project Dependable Internet of Things funded by Graz University of Technology and the SiliconAlps project Archimedes funded by the Austrian Research Promotion Agency (FFG). 
This project has further received funding from the European Union'€™s Horizon 2020 research and innovation programme under the Marie Sk\l{}odowska-Curie Grant Agreement No.~797223 --- HYBSPN.

We acknowledge NVIDIA for providing GPU computing resources.


\begin{IEEEbiography}{Franz Pernkopf}
received his MSc (Dipl. Ing.) degree in Electrical Engineering at Graz University of Technology, Austria, in summer
1999. He earned a PhD degree from the University of Leoben, Austria, in 2002. In 2002 he was awarded the Erwin Schr\"{o}dinger Fellowship. He was a Research Associate in the Department of Electrical Engineering at the University of Washington, Seattle, from
2004 to 2006. Currently, he is Associate Professor at the Laboratory of Signal Processing and Speech Communication, Graz University
of Technology, Austria. His research interests include machine learning, resource-efficient neural networks, probabilistic graphical models, and speech processing applications.
\end{IEEEbiography}

\begin{IEEEbiography}{Wolfgang Roth} 
received the BSc and MSc degrees (with distinction) in computer science from Graz University of Technology in 2015. Currently, he is with the Signal Processing and Speech Communication Laboratory at Graz University of Technology where he is working toward the PhD degree. His research interests include resource-efficient deep neural network, Bayesian learning and statistical pattern recognition.
\end{IEEEbiography}

\begin{IEEEbiography}{Matthias Z\"ohrer}
received his MSc (Dipl. Ing.) degree in Telematik at Graz University of Technology, Austria, in summer
2013. Since 2013 he is a Research Associate at the Laboratory of Signal Processing and Speech Communication, Graz University of Technology, Austria. His research interests include deep learning, GPU optimized processing, robust speech recognition, speech enhancement and beamforming.
\end{IEEEbiography}

\begin{IEEEbiography}{Lukas Pfeifenberger}
received the M.Sc. (Dipl. Ing. FH) degree in computer science from the University of Applied Sciences, Salzburg, Austria, in 2004. Since 2005 he has been working in the electronics industry on projects pertaining to FPGA design, DSP programming and communication acoustics. In 2013, he received the M.Sc. (Dipl. Ing.) degree in Telematik at Graz University of Technology, Austria. Since 2015 he has been a Research Associate at the Laboratory of Signal Processing and Speech Communication, Graz University of Technology, Austria. His research interests include machine learning, computer vision and speech enhancement. He currently pursues research projects in blind source separation and acoustic echo control.
\end{IEEEbiography}

\begin{IEEEbiography}{G\"unther Schindler}
is a doctoral candidate at the Institute of Computer Engineering at the Ruprecht-Karls University of Heidelberg (Germany). He has received his MSc in computer engineering and BSc in electrical engineering in 2016 from the Ruprecht-Karls University of Heidelberg respectively in 2014 from University of Applied Sciences Munich. His research interests include architectures and algorithms for application specific computing on accelerators and methods for resource-efficient deep neural networks.
\end{IEEEbiography}

\begin{IEEEbiography}{Holger Fr\"oning}
is since 2011 an associate professor at the Institute of Computer Engineering at the Ruprecht-Karls University of Heidelberg (Germany). During this time, he was as visiting scientist with NVIDIA Research (Santa Clara, CA, US), and visiting professor at Graz University of Technology, Austria. From 2008 to 2011 he reported as postdoctoral fellow to Jose Duato from Technical University of Valencia (Spain). He obtained his PhD and MSc degrees 2007 respectively 2001 from the University of Mannheim, Germany. He has received a Google Faculty Research Award in 2014, and best paper awards at ICPP and IPDPS. His research interests include hardware and software architectures, related co-design, data movement optimizations, and power and energy consumption of computing systems. 
\end{IEEEbiography}

\begin{IEEEbiography}{Sebastian Tschiatschek} 
received the BSc and MSc degrees (with distinction) in electrical
engineering from the Graz University of Technology (TUG) in 2007 and 2010, respectively.
He conducted his master's thesis during a one-year stay at ETH Zurich, Switzerland. In 2014 he earned a PhD degree at the Signal Processing and Speech Communication Laboratory at TUG. From 2015-2017 he was a postdoc in the Learning and Adaptive Systems Group at ETH Zurich. Currently he is a research associate at Microsoft Research Cambridge, UK. His research interests include submodular functions, their 
application in the machine learning related applications, Bayesian networks and inference under computational constraints.
\end{IEEEbiography}

\begin{IEEEbiography}{Robert Peharz} 
received the MSc degree in computer engineering and the PhD degree in electrical engineering from Graz University of Technology.
His main research interest lies in machine learning, in particular probabilistic modeling, with applications to signal processing, speech and audio processing, and computer vision. 
Currently, he is a Postdoc in the Machine Learning Group at the University of Cambridge.
He was recently awarded with a Marie-Curie Individual Fellowship.
\end{IEEEbiography}

\begin{IEEEbiography}{Matthew Mattina}
is Senior Director of Machine Learning \& AI Research at Arm, where he leads a team of researchers developing advanced hardware, software, and algorithms for machine learning. Prior to joining Arm in 2015, Matt was Chief Technology Officer at Tilera, responsible for overall company technology, architecture and strategy. Prior to Tilera, Matt was a CPU architect at Intel and invented and designed the Intel Ring Uncore Architecture. Matt has been granted over 30 patents relating to CPU design, multicore processors, on-chip interconnects, and cache coherence protocols. Matt holds a BS in Computer and Systems Engineering from Rensselaer Polytechnic Institute and an MS in Electrical Engineering from Princeton University.
\end{IEEEbiography}

\begin{IEEEbiography}{Zoubin Ghahramani} 
Zoubin Ghahramani is Professor of Information Engineering at 
the University of Cambridge, where he leads the Machine Learning Group. 
He studied computer science and cognitive science
at the University of Pennsylvania, obtained his
PhD from MIT in 1995, and was a postdoctoral
fellow at the University of Toronto. 
His academic career includes concurrent appointments as one
of the founding members of the Gatsby Computational Neuroscience Unit in London, 
and as a faculty member of CMU's Machine Learning
Department for over 10 years. 
His research interests include statistical machine learning, Bayesian
nonparametrics, scalable inference, probabilistic programming, and building an automatic
statistician. 
He has held a number of leadership roles as programme
and general chair of the leading international conferences
in machine learning: AISTATS (2005), ICML (2007, 2011), and NIPS (2013, 2014).
In 2015 he was elected a Fellow of the Royal Society.
\end{IEEEbiography}

\clearpage

\end{document}